\documentclass{article}
\usepackage{PRIMEarxiv}
\pdfoutput=1
\usepackage[utf8]{inputenc} 
\usepackage[T1]{fontenc}    
\usepackage{hyperref}       
\usepackage{url}            
\usepackage{booktabs}       
\usepackage{amsfonts}       
\usepackage{nicefrac}       
\usepackage{microtype}      
\usepackage{fancyhdr}       
\usepackage{graphicx}       

\usepackage{enumitem}
\usepackage{subcaption}
\usepackage{dsfont}
\usepackage{multirow}
\usepackage{amsmath}
\usepackage{bm}
\usepackage{url}

\usepackage[style=authoryear-ibid, backend=biber, uniquename=false, maxcitenames=2, uniquelist=false]{biblatex}
\let\cite\parencite
\citetrackerfalse
\title{Towards Error Measures which Influence a Learners Inductive Bias to the Ground Truth
}

\author{A.\ I.\ Parkes \\
       University of Southampton\\
       \texttt{A.I.Parkes@soton.ac.uk} \\
       \And
       A.\ J.\ Sobey \\
       University of Southampton\\
       The Alan Turing Institute\\
       \texttt{ajs502@soton.ac.uk}
       \And
       D. A. Hudson \\
       University of Southampton\\
       \texttt{dominic@soton.ac.uk} \\
  
}

\begin{document}
\maketitle

\begin{abstract}
Artificial intelligence is applied in a range of sectors, and is relied upon for decisions requiring a high level of trust. For regression methods, trust is increased if they approximate the true input-output relationships and perform accurately outside the bounds of the training data. But often performance off-test-set is poor, especially when data is sparse. This is because the conditional average, which in many scenarios is a good approximation of the `ground truth', is only modelled with conventional \mbox{Minkowski-r} error measures when the data set adheres to restrictive assumptions,  with many real data sets violating these. To combat this there are several methods that use prior knowledge to approximate the `ground truth'. However, prior knowledge is not always available, and this paper investigates how error measures affect the ability for a regression method to model the `ground truth' in these scenarios. Current error measures are shown to create an unhelpful bias and a new error measure is derived which does not exhibit this behaviour. This is tested on 36 representative data sets with different characteristics, showing that it is more consistent in determining the `ground truth' and in giving improved predictions in regions beyond the range of the training data.
\end{abstract}


\section{The Importance of Correctly Interpreting the Ground Truth}
The artificial intelligence industry is expected to grow to \$40 billion dollars by 2025 \cite{Herald}, revolutionising sectors from manufacturing to healthcare. Increasing the use of artificial intelligence for applications where dependability, safety or security are of concern requires trustworthy and explainable methods to facilitate human understanding and decision making. Regression problems, where relationships between the inputs and continuous output(s) are identified, comprise a large portion of the cutting edge of the machine learning field \cite{li2020}. However, none of the commonly used performance error measures for regression problems assess whether the input-output relationships are modelled accurately; methods reporting low prediction error do not necessarily approximate the ground truth. If it cannot be verified that these methods accurately approximate the ground truth of their data sets then they cannot be trusted to perform in many real-world applications.

There are many situations where regression is required but there is not a full understanding of the input-output relationships, these range from engineering \cite{Parkes2018} \cite{richmond2020} and financial \cite{prado2018} to medical applications \cite{sidey2019} \cite{wang2010}. This can be due to a lack of domain knowledge as well as sparse or unrepresentative data sets, because data can be expensive to gather in large quantities or because the phenomenon of interest is rare. Machine learning methods like neural networks have large modelling flexibility, given sufficient data they can find patterns in problems where complexity prohibits the explicit programming of a systems precise physical nature. But they are known to produce physically inconsistent results and cannot generalise to off test sets, so have therefore not modelled the ground truth of the system \cite{willard2020}. 

\begin{figure}[h]
\centering
    \includegraphics[width=0.5\linewidth]{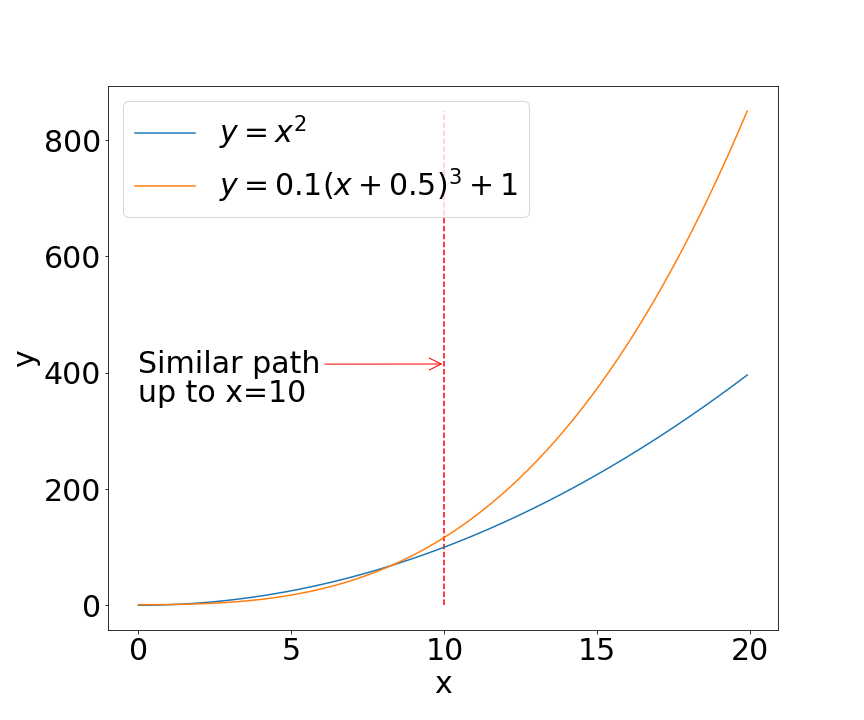}
\caption{Example of two curves tracking similar patterns in the x range from 0-10 but diverging above x values of 10.}\label{fig:lines}
\end{figure}

Due to the flexibility of machine learning regression methods, any number of arbitrary patterns could be modelled which provide a good approximation to the fundamental relationship over a limited input-output domain, despite these functions having no similarities outside of this range; this effect is illustrated for two functions in Figure \ref{fig:lines}. This demonstrates the common issue of poor extrapolation for current machine learning techniques. For example, a method trained on the $x$ domain between $0-10$, with limited or poor quality data for the $x$ domain above 10, would have no way to discern which of the two curves was the true $x-y$ relationship. Both curves would produce similar error profiles for $x<10$, using current error measures.


Therefore it cannot  be assumed that regression methods producing low \mbox{Minkowski-r} error measures, such as the Mean Squared or Mean Absolute Errors, model the ground truth accurately. Even methods to balance the bias-variance trade-off, such as regularisation and cross-validation, act to minimise the error or to fix overfitting but don’t account for the ground truth. In fact, there are restrictive assumptions about a data set that must be met to ensure that a regression method approximates the conditional average of the data set, the average output value conditioned on each input in turn\cite{Bishop1995}. One of the methods with the least restrictive assumptions are neural networks, and these assumptions are discussed further in Section \ref{minkowski} where it is noted that the number of applications which violate these assumptions is larger than is apparent in the standard literature. 


Even if there is certainty that the regression producing minimal loss approximates the conditional averages of the data set, the conditional averages do not necessarily approximate the fundamental relationships which generated the data set. This can be caused by sparse areas of data which can be affected by the `law of small numbers' creating misleading averages \cite{Tversky71}. Additionally, the interaction of noise and high convexity or concavity in input-output relationships creates a gap between the conditional averages of a data set and the underlying relationships which generated them, a direct result from Jensen's inequality \cite{Jensen}, discussed in more detail in Section \ref{error measure}.

To address this problem a number of methods exist to improve a method’s approximations of the ground truth. These methods include biasing the architecture of learning methods to known relationships \cite{Park2019} \cite{Anderson2019}\cite{Zhang2018}; using physics-guided initialization \cite{read2019}\cite{sultan2018}; and adding a `distance from the ground truth' measure to the loss function \cite{Karpatne2018}. They can be used to produce Reduced Order Models \cite{lucia2004} or improve the prediction from the physical model alone. These have been shown to produce better extrapolation predictions as their approximated patterns are robust to sparse regions of data \cite{willard2020}. Methods also exist to mitigate poor inductive biases or learning biases, where a method tends towards error minima which produce trained methods that do not generalise off-test set, by generating extra data \cite{abusitta2020}. However, as the discussed methods either train models so that they have an inbuilt bias to the known input-output relationships, or generate extra data based on an understanding of the system being modelled, knowledge of these relationships is required. There is currently no way to measure the extent to which a method approximates the ground truth if it is not already known.

All regression methods require a loss function to produce predictions on a test set. Traditional maximum likelihood regression approaches such as generalized linear models, support vector regressions, and other kernel regressions require an intelligent selection of multiple model parameters \cite{nelder1972} and \cite{drucker1996}. These parameter choices are often based on variable distributions, variable dependence and noise characteristics \cite{murphy2012} which are not necessarily known. Bayesian methods, including Gaussian processes and Bayesian neural networks, provide a solution to not knowing the best parameter value, or basis function, to use in a model by using `non-informative priors'\cite{bernardo1979}, and integrating over all feasible values to identify the optimal parameters for any given application\cite{lampinen2001}. However, in addition to the increased computational complexity this causes\cite{green2015}, applications of the `no free lunch' theorem state that if you make no assumptions concerning the target, then you have no assurances about how well you generalise \cite{mackay1992b} so if there is no prior knowledge to build into a Bayesian method and `non-informative priors' are used exclusively, then the model will have limited practical uses. In addition, despite being able to perform inference without reference to a loss function, Gaussian processes still require a loss function to identify the optimal prediction for a testing set \cite{rasmussen2006}. They use the same \mbox{Minkowski-r} family of error measures and therefore have the same problems determining the conditional average.  

This paper focuses on the situation where there is a partial, or no, understanding of the input-output relationships in a data set, so current approaches to fitting to the ground truth can not be used. A new error measure to assess how accurately the ground truth is mapped by a regression is proposed and is shown to be effective. The interaction between the conditional averages and the ground truth is also formalised for these data sets. It highlights that in these situations minimising traditional error measures cannot guarantee an accurate approximation of the ground truth, and discusses potential causes for the poor inductive bias produced by these error measures. The error measures developed in this study can be applied to any regression method, as they all require the use of a loss function to make predictions. Neural networks are employed as the exemplary method in this study as they are used prolifically in the machine learning literature due to their flexibility and ease of implementation. 

\section{Current Regression Error Measures}\label{minkowski}

Nearly all common regression error measures are point-based and can be defined as in \cite{botchkarev2018}:

\begin{equation*}
\mathds{G}_{i=1,\dots,n}(\mathds{N}[\mathds{D}(y_{i},\hat{y_{i}})]), \label{eq:general}
\end{equation*}

\noindent where $\mathds{D}$ is some distance measure between the two points $\hat{y_{i}}$ the predicted target variable(s) and $y_{i}$ from the measured data set, $\mathds{N}$ is some normalisation method, and $\mathds{G}$ some aggregation method. Other error measures include the $R^{2}$ or Pearson correlation coefficient but interpreting model performance from correlation measures is more difficult than interpreting point-based error measures \cite{taylor1990}, as they provide limited intuition about model behaviour for a single prediction or use in `real-life' scenarios.

The most popular error measures used for regression are the \mbox{Minkowski-r} distance measures \cite{Hanson1987}, where the distance measure is the Euclidean distance to some power $r$ and the aggregation is the mean,

\begin{equation*}
M = \frac{1}{r} \sum_{i} \mathds{N}[(|y_{i}-\hat{y_{i}}|)^{r}], r\in \mathds{R}. \label{eq:Mr}
\end{equation*}

In these error measures higher $r$ values increase the penalty for large deviations, and smaller $r$ values reduce the influence of outliers in feature space during learning. When the regression output only has one dimension, these measures equate to Mean Squared Error for $r=2$ and Mean Absolute Error for $r=1$. Other variants can be produced by varying the normalisation method; for example, Mean Absolute Percentage Error is produced using a normalisation with the target variable value. Theoretically, minimising the Mean Squared Error of a method creates an approximation of the conditional mean of the data set \cite{Bishop1995}, and a method which minimises Mean Absolute Error should approximate the conditional median of the data, as this is less affected by outliers, however, these proofs rely heavily on several assumptions. The assumptions vary slightly for different methods, one of the least restrictive sets is that for neural networks:

\begin{enumerate}[label=(\roman*)]
\item the datapoints are independent; 
\item the distribution of the target variable is to be deterministic of the input with Gaussian noise, e.g. $y=\phi(x)+\epsilon$ where $\phi$ depends only on input variable x, and $\epsilon \sim N(0,\sigma^{2})$; 
\item the standard deviation of noise, $\sigma$, is not dependent on the input x; 
\item and the data set and neural network must be sufficiently large. 
\end{enumerate}

These are not trivial assumptions; requiring that the standard deviation of the noise is not dependent on the input values implies that only homoscedastic data sets can be modelled. In many applications these assumptions do not hold, although meaningful results can still be produced by applying neural networks with Mean Squared Error or Mean Absolute Error error measure there is no certainty that the trained networks approximate the conditional average of the data set. For example, if noise in the target variable is not Gaussian, then the results cannot distinguish between the true distribution and any other distribution with the same mean and variance. It is suggested that this is the reason regression networks can model a data set with low errors, but model arbitrary patterns rather than the conditional average of the data set. 

This paper specifically investigates data sets where at least one $\phi$, the input-output relationship for a single input feature, is a non-linear function. So there is some curvature, either concave or convex, in the system being modelled. For applications modelling physical causality in a system a conditional average is likely to approximate the ground truth of the data set, although depending on the characteristics of input-output relationship $\phi$ and any noise the choice of conditional average may be important. By synthesising results from \cite{Jensen} and \cite{mean_median_mode} we have that, for data sets where $\phi$ has a larger curvature and higher standard deviations of noise, the conditional median will approximate the ground truth closer than the conditional mean compared to data sets with a lower curvature and smaller standard deviation of noise, where the mean and median will tend to the same value as the conditional distributions have less of a skew. 

\begin{figure}[h]
\centering
\subfloat[]{\includegraphics[width=0.3\linewidth]{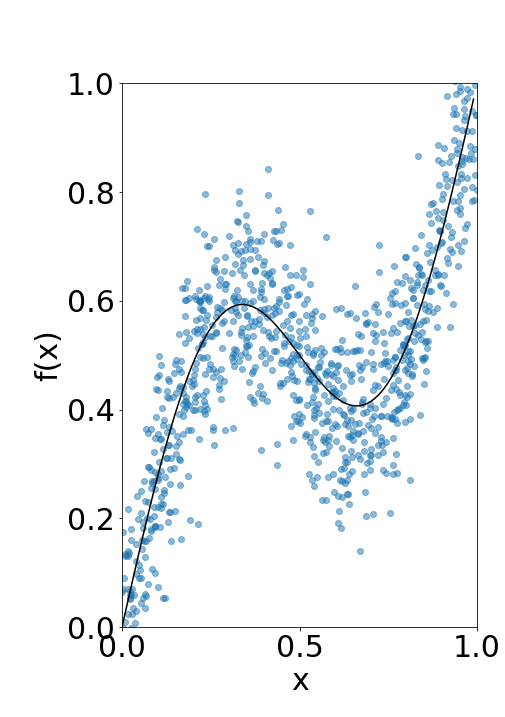}
\label{fig:inv}}
\subfloat[]{\includegraphics[width=0.3\linewidth]{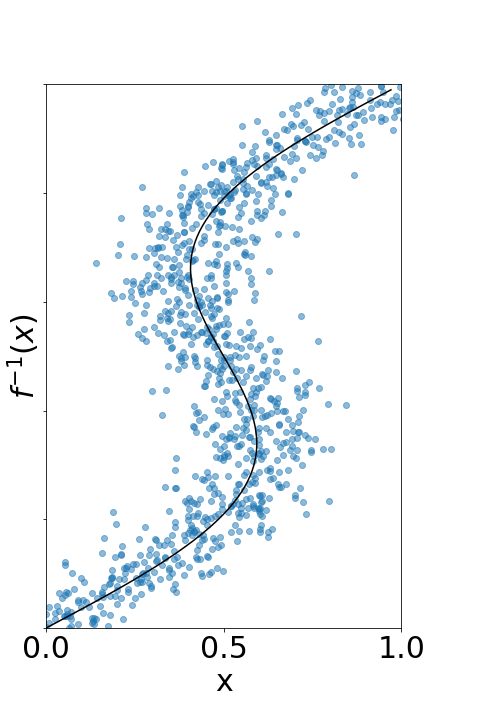}
\label{fig:invinv}}
\caption{Illustrations of an inverse problem, where $f(x) = x + 0.3 sin(2\pi x) + \epsilon$ with $\epsilon \sim N(0,0.1)$. A) $f(x)$ against $x$, where $f$ is a function also showing the ground truth without noise and B) the inverse, $f^{-1}(x)$ which is not a function, where the conditional average of $f^{-1}(x)$ does not approximate the ground truth relationship $f^{-1}$.}\label{fig:inverse}
\end{figure}

However, the conditional average does not necessarily approximate the ground truth for all data sets. For example if the relationship between inputs and outputs are a relation and not a function, there are several valid values for the outputs, illustrated in Figure \ref{fig:inverse}. This means the conditional average of the data set can hold no useful information about the data at all, as the average of several solutions is not necessarily itself a solution.

The use of \mbox{Minkowski-r} error measures, most commonly Mean Squared Error and Mean Absolute Error, on data sets which do not adhere to the above assumptions leads to poor performance outside areas of dense data as the ground truth has not been approximated. This reduces the level of trust a user has for a model and inhibits our understanding of why certain predictions are correct and some are incorrect. This lack of reproducibility is a growing problem in the field, \cite{Hutson2018} and our lack of understanding means that these models are unusable in some domains \cite{Voosen2017}. It also reduces the transferability of trained methods as the causal relationships are not identified by these models.

\section{Determination of a New Error Measure: Mean Fit to Median}\label{error measure}

To derive an error measure which quantifies how accurately a regression method models the  ground truth in a data set, an approximation of the isolated input-output relationships is required. Therefore, the proposed new measure calculates a proxy for the relationship between each input variable and the output, the conditional average independent of other input variables. It measures the distance between the approximation of these relationships and the proxy curves. 

First, these relationships are formalised for any data set; for clarity the principles of the new error measure are explained using a two dimensional example, where input $x$ is used to predict output $y$, and this is expanded to include scenarios with multi-dimensional inputs for the formal statement of the error measure. In this initial reduced set-up, let $\phi$ be the relationship between input variable $x$ and output variable $y$, such that $y=\phi(x)$. For simplicity, the only curves $\phi$ discussed are either convex or concave across their entire domain, although it is suggested that the arguments extend to a piecewise curve.

\begin{figure}[h]
\centering
\subfloat[]{\includegraphics[width=0.5\linewidth]{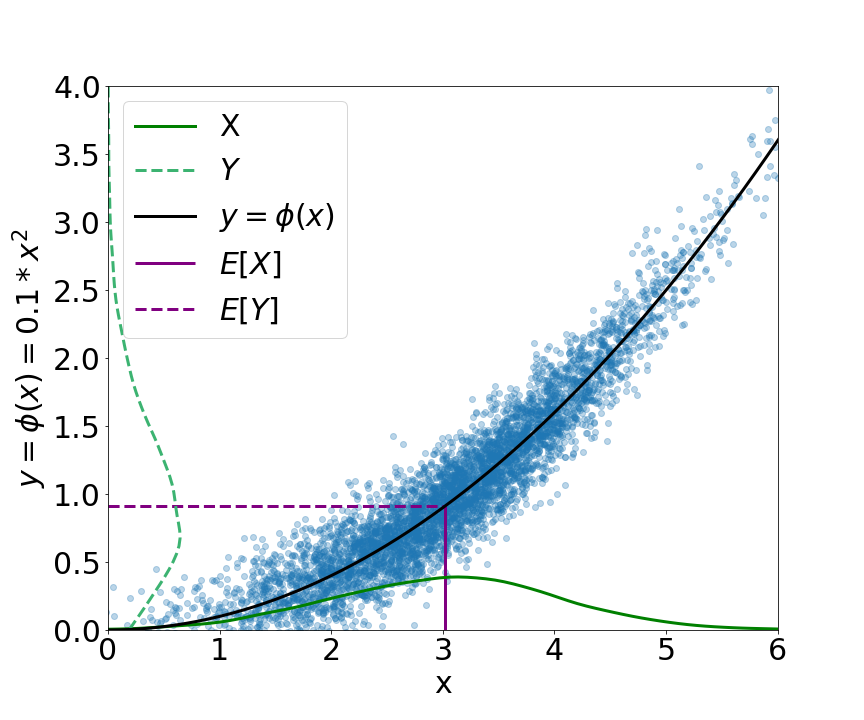}
\label{fig:setup}}
\subfloat[]{\includegraphics[width=0.5\linewidth]{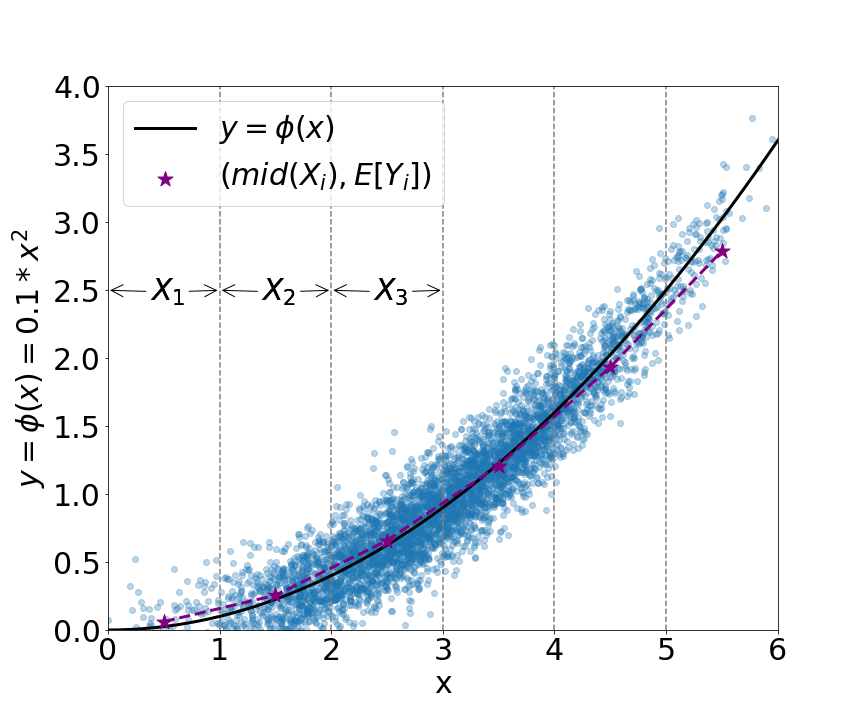}
\label{fig:partition}}
\caption{Illustration of (A) expectation of X and Y distribution, (B) partitioning X distribution to create proxy curve.}\label{fig:illustrate}
\end{figure}

From this data set we have the distributions $X$ and $Y(=\phi(X))$ shown on the axis of Figure \ref{fig:setup} and $E[X]$ and $E[Y](=E[\phi (X)])$ which are defined as the expectation, or mean, of the distributions. Assuming these distributions are sufficiently continuous, a naive proxy for $\phi$ can be derived by estimating the conditional averages of the data set. To produce this the $X$ domain must first be divided into $n$ equally spaced bins ($X_{1},X_{2},\dots,X_{n}$), Figure \ref{fig:partition}. Then $Y (=\phi (X))$ is  partitioned based on the data points' corresponding $x$ values, producing ($Y_{1},Y_{2},\dots,Y_{n}$).  $Y_{i}$ is therefore the set of all the observed output values for the  input values in $X_{i}$. Assuming  a sufficiently large set of partitions $n$, the mean of all $Y_{i}$ and the midpoints of $X_{i}$ create an initial proxy for the conditional average: (mid($X_{i}), E[Y_{i}])$ for $i=1,2,\dots,n$. 

\begin{figure}[h]
\centering
\subfloat[]{\includegraphics[width=0.5\linewidth]{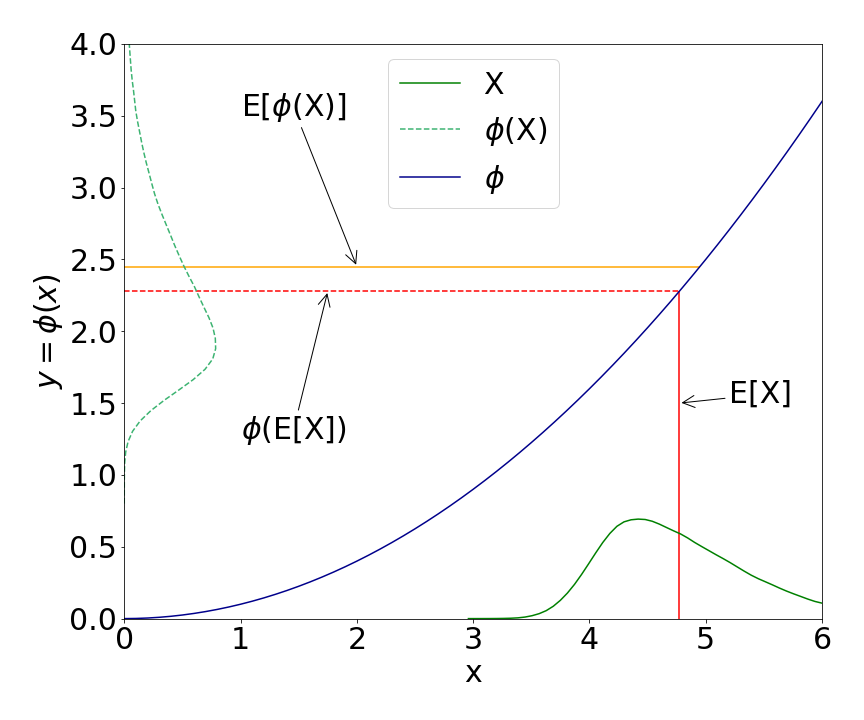}
\label{fig:jensen_vanilla}}
\subfloat[]{\includegraphics[width=0.5\linewidth]{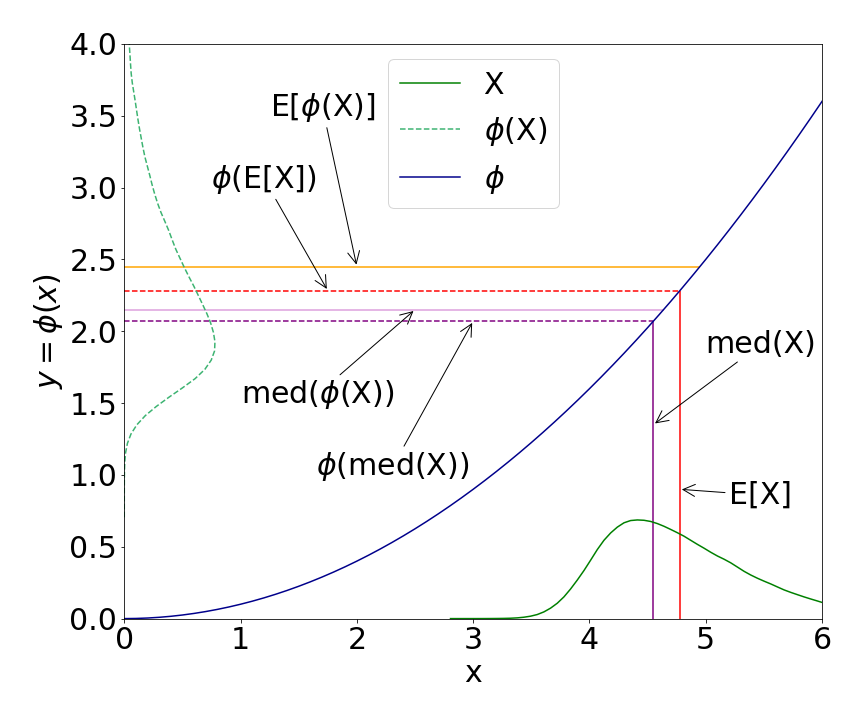}
\label{fig:jensen_medians}}
\caption{Illustrations of Jensen's inequality $\phi (E[X]) \leq E[\phi (X)]$ for convex function $\phi$ where A) the transformed mean of a distribution $\phi(E[X])$ is less than the mean of the transformed distribution $E[\phi(X)]$ \cite{Jensen} and B) where the gap caused by this inequality is smaller for the median of the distributions than for the mean. \cite{Jen_exten}.}\label{fig:jensen}
\end{figure}

However, from Jensen’s inequality we have that $\phi (E[X]) \leq E[\phi (X)]$ for convex\footnote{And  $\phi (E[X]) \geq E[\phi (X)]$ for concave $\phi$.} $\phi$ \cite{Jensen}. This is illustrated in Figure \ref{fig:jensen_vanilla} where the function value at the average of the input variable distribution ($\phi(E[X])$) is less than the average of the distribution of the function values of the inputs ($E[\phi (X)]$). This means that for any $\phi$ which is not a straight line, a gap between $\phi (E[X]) $ and $ E[\phi (X)]$ will exist and hence the mean of $Y_{i}$ will not approximate $\phi (X_{i})$ for these applications. Formal bounds on this inequality exist \cite{bound_Jen}, but cannot be applied without prior knowledge of $\phi$.  

We know that $\text{med}(\phi (X)) \leq E[\phi (X)]$ for unimodal right-skewed distributions from basic statistics \cite{mean_median_mode}, so the median of each $Y_{i}$ (denoted $\text{med}(\phi (Y_{i}))$) could produce better estimates of $\phi$ under certain constraints. If $\phi$ is non-linear then $X_{i}$ and $Y_{i}$ will have a skew and, as continuity is already assumed, these subsets of the distributions will be unimodal. Jensen's inequality also holds for medians; $\phi (\text{med}(X)) \leq \text{med}(\phi (X))$ for convex $\phi$ \cite{Jen_exten}, Figure\ref{fig:jensen_medians}. So although there will also be a gap between $\text{med}(\phi (X))$ and the true curve  $\phi$ , $\text{med}(Y_{i})$ will produce a closer approximation than $E[Y_{i}]$. The proposed measure is therefore the `Mean Fit to Median' which is the mean of the distance between the proxy points for the conditional median and the isolated input-output relationships predicted by a machine learning regression method. 

Here the notation is expanded to include multiple input variables, so that there are $m$ independent input variables $x^{j}$ such that $y = \phi(x^{1}, x^{2}, \dots , x^{m})$. The distribution of each input variable is partitioned as above, formalised this produces

\begin{equation*}
    \text{proxy curves} = \{\text{med}(Y^{j}_{i}) | i=1,2, \dots, n, j=1,2,\dots,m\},
\end{equation*}

\noindent where $Y^{j}_{i}$ is the set of $y$ values corresponding to the set $X^{j}_{i}$. 

The distance between these proxy curves and the input-output relationships learnt during training is calculated to produce the Mean Fit to Median Error value. The input-output relationships learnt by a trained regression method, are produced by probing a trained method to predict output values for each input variable in turn by using inputs, cycling from the minimum to maximum observed values of each input, with all other inputs at the median value. Explicitly, if $P$ denotes the representations learnt from a model such that $P(x^{1},x^{2},\dots,x^{m})$ is the predicted output value for inputs $x^{1},x^{2},\dots,x^{m}$. Then for input variable $j$, the predicted input-output relationship is approximated by the set

\begin{equation*}
    \text{learnt input-output curves} = \{p^{j}_{X_{i}} | i=1,2, \dots, n, j=1,2,\dots,m\},
\end{equation*}

\noindent where $p^{j}_{X_{i}} = P\left(\text{mid}(X^{j}_{i}), \text{med}(X^{k}) \text{ for } k\neq j\right)$ with mid($X$) denoting the midpoint of the set $X$. This allows the Mean Fit to Median Error measure to be written as

\begin{equation}
        \text{Mean Fit to Median} = \frac{1}{m} \sum^{m}_{j=1} \frac{1}{n} \sum^{n}_{i=1} \left|\text{med}(Y^{j}_{i}) - p^{j}_{X_{i}}\right|.
        \label{eq:MFTM}
\end{equation}

It is only possible to calculate this error measure after normal training has terminated, as the input-output relationships learnt by the method are required to perform the calculation. This means that currently it is only possible to use the `Fit to Median' measure as a post-training evaluation method. However, it can be used to identify the method with the best ground truth approximation from a set of trained methods. Due to the extension of Jensen's inequality to medians\cite{Jen_exten} the Fit to Median does not generalise to solve inverse problems, as illustrated in Figure \ref{fig:inverse}.

\section{Generating Artificial Data Sets}\label{data}
For many real regression applications a full understanding of the system and its uncertainties is not available. This means that benchmarking to what extent a trained method replicates the true relationships between inputs and outputs is not possible. To allow this to be accurately measured artificial data sets are used, with a range of fully defined variable relationships and differing levels of uncertainty, with the aim of approximating the complexity and characteristics of common regression problems. The data is also generated with the intention of violating assumptions (ii)-(iv), discussed in Section \ref{minkowski}.

This data generation process is formalised as follows: the input variables are generated such that  $\bm{X} \sim N_{6}(\bm{\mu},\bm{\Sigma})$,  with $\bm{\mu} \in \left\{(\mu_{0},\dotsc , \mu_{5}) \middle| \mu \in \mathds{R}\right\}$ and for initial benchmarking $\bm{\Sigma}$, the covariance matrix, is diagonal with a view to investigate scenarios where input variables are not independent in a further study. In practice $\mu_{i}$ is randomly generated as $\mu_{i} \sim \text{unif}(0,10)$ from numpy random number modules and the number of observations generated is 1,000,000. To generate one output $y$ the following procedure is used; 
\begin{equation*}
 y = \phi(x_{0}, x_{1}, \dots, x_{5}) = \phi(\bm{X}) = \bm{W \cdot F}(\bm{X}),
\end{equation*}
where $\bm{W} \in \left\{ (w_{0}, \dotsc , w_{5}) \middle| w_{i} \in [0,1] , \sum w_{i} = 1 \right\}$ and,
\begin{equation*}
\bm{F} = (\tilde{f}_{0} , \dotsc , \tilde{f}_{5}).
\end{equation*}
Where noise in the $y$ direction is introduced as 
\begin{equation*}
\tilde{f}(x) = f(x) + \epsilon,
\end{equation*}
with
\begin{equation*}
\epsilon \sim N(0, \sigma ^{2}),
\end{equation*}

and $\sigma \in \mathds{R}$ not dependent on x. The isolated input-output functions, $f_{i}$ are represented as 
\begin{equation*}\label{eq:funcs}
f \in \mathds{P}_{5}[x],
\end{equation*}

where $ \mathds{P}_{n}[x] = \left\{ p_{0} + p_{1}x + p_{2}x^{2} + \dotsb + p_{n}x^{n} \middle| p_{i} \in \mathds{R} \right\}$. For each data set the degree of the polynomials  $f_{i}$ are randomly chosen from the range specified, for example $\mathds{P}_{a-b}$ indicates the degrees of $f_{i}$ are chosen integers from the range from $a-b$ inclusive. Then the coefficients are randomly selected uniformly from the interval $(-10,10)$.

\begin{figure}[h]
\centering
\subfloat[]{\includegraphics[width=0.55\linewidth]{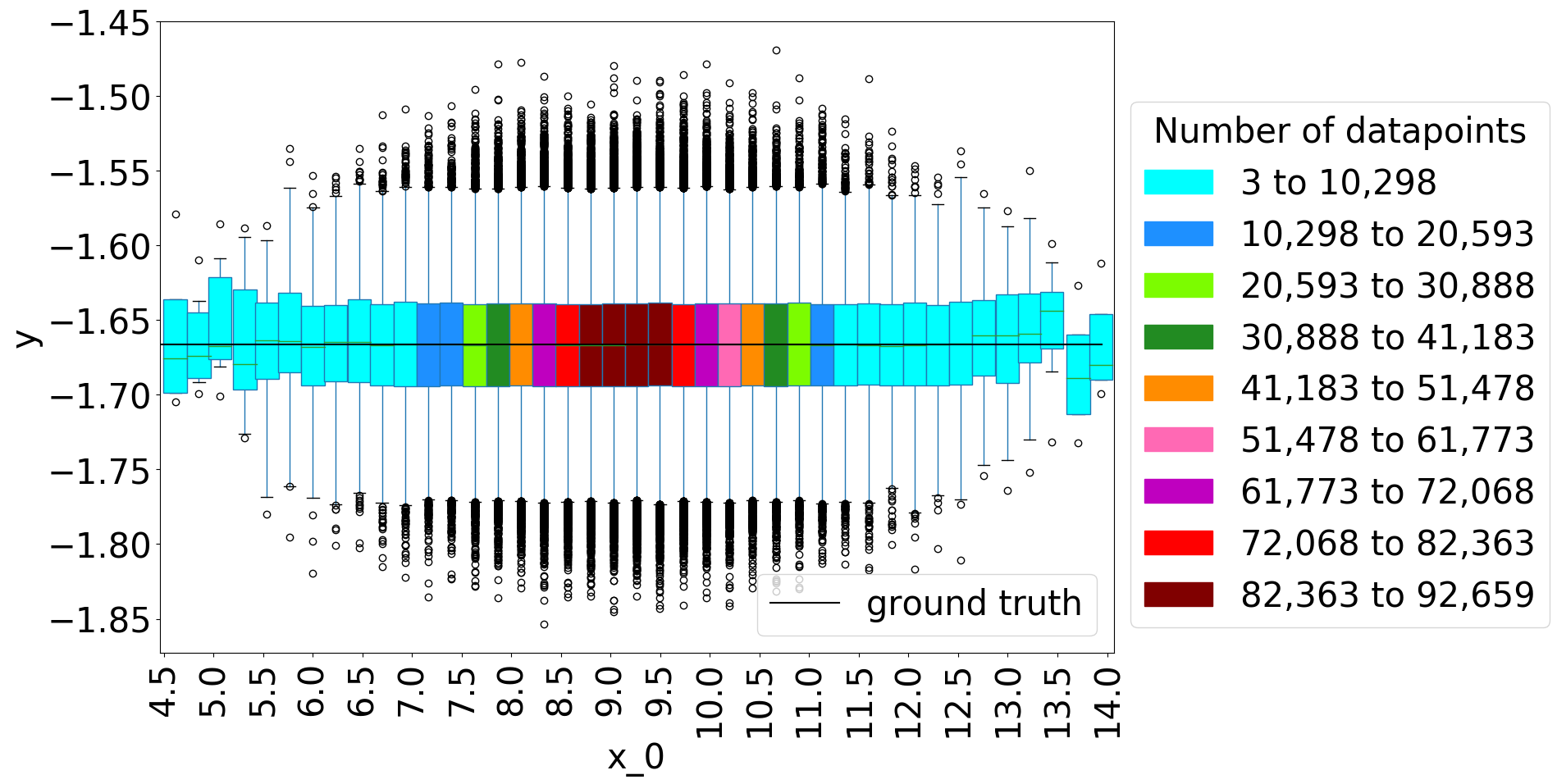}
\label{fig:box_0}}
\subfloat[]{\includegraphics[width=0.4\linewidth]{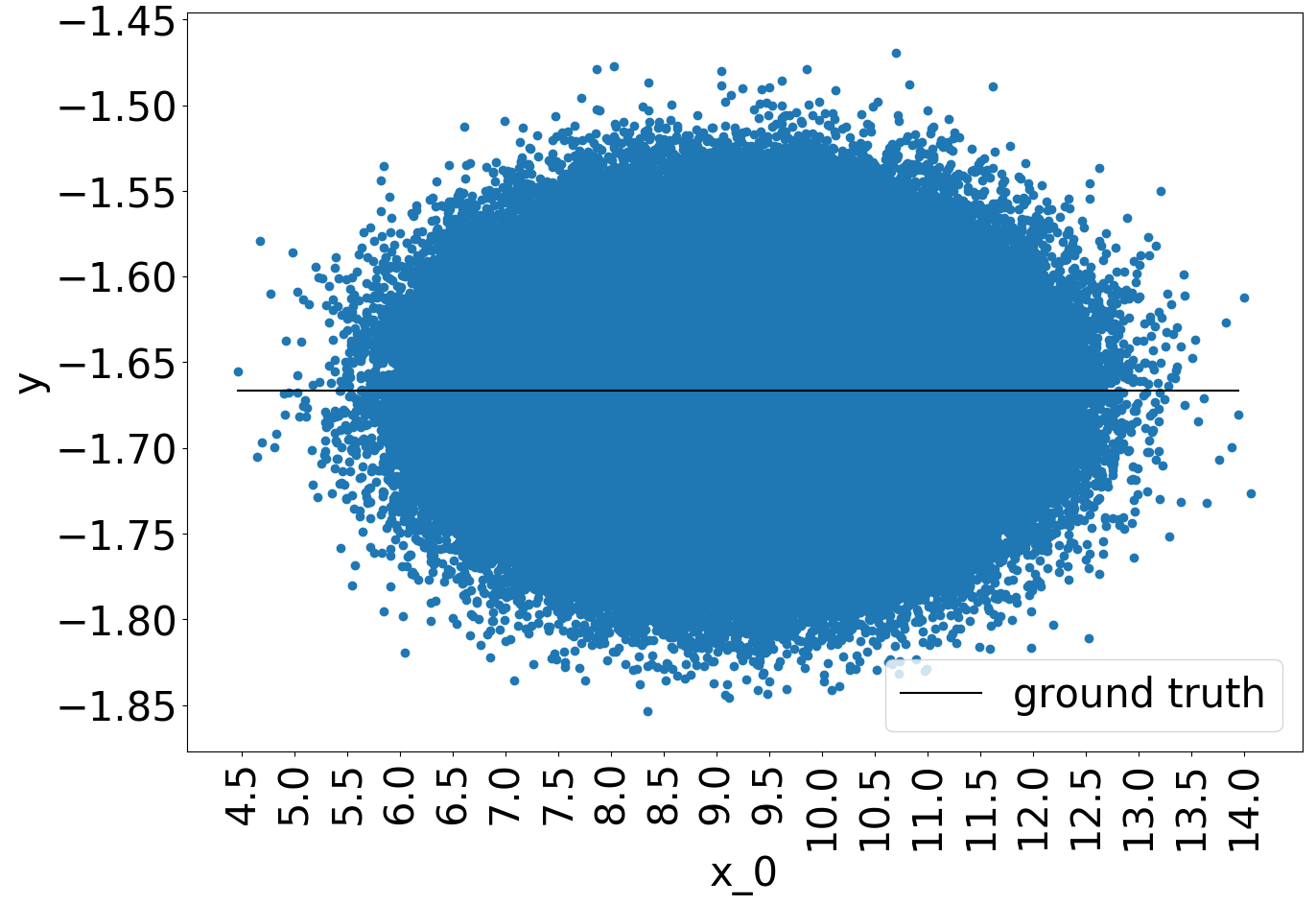}
\label{fig:scatter_0}}\\
\subfloat[]{\includegraphics[width=0.55\linewidth]{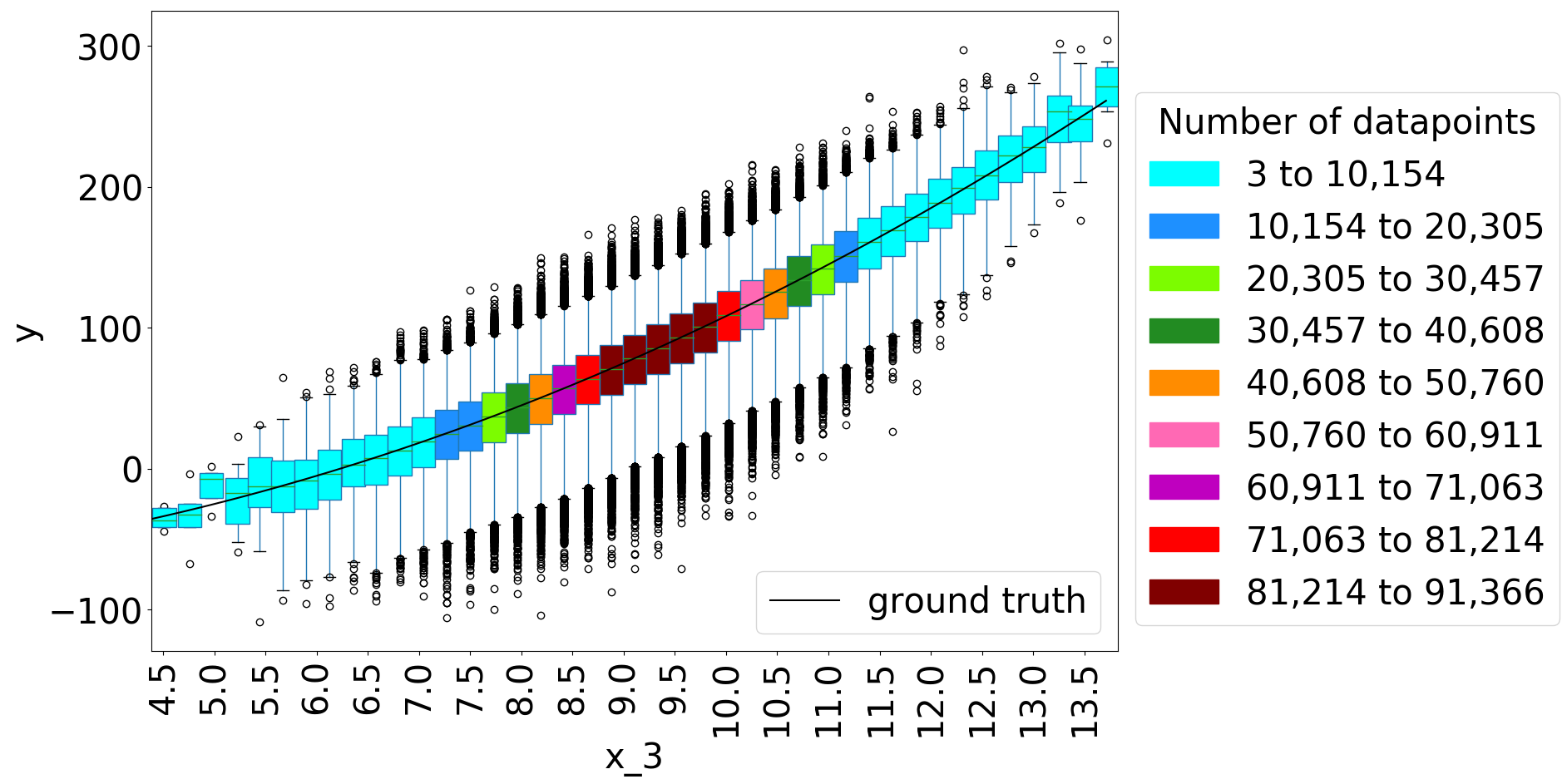}
\label{fig:box_2}}
\subfloat[]{\includegraphics[width=0.4\linewidth]{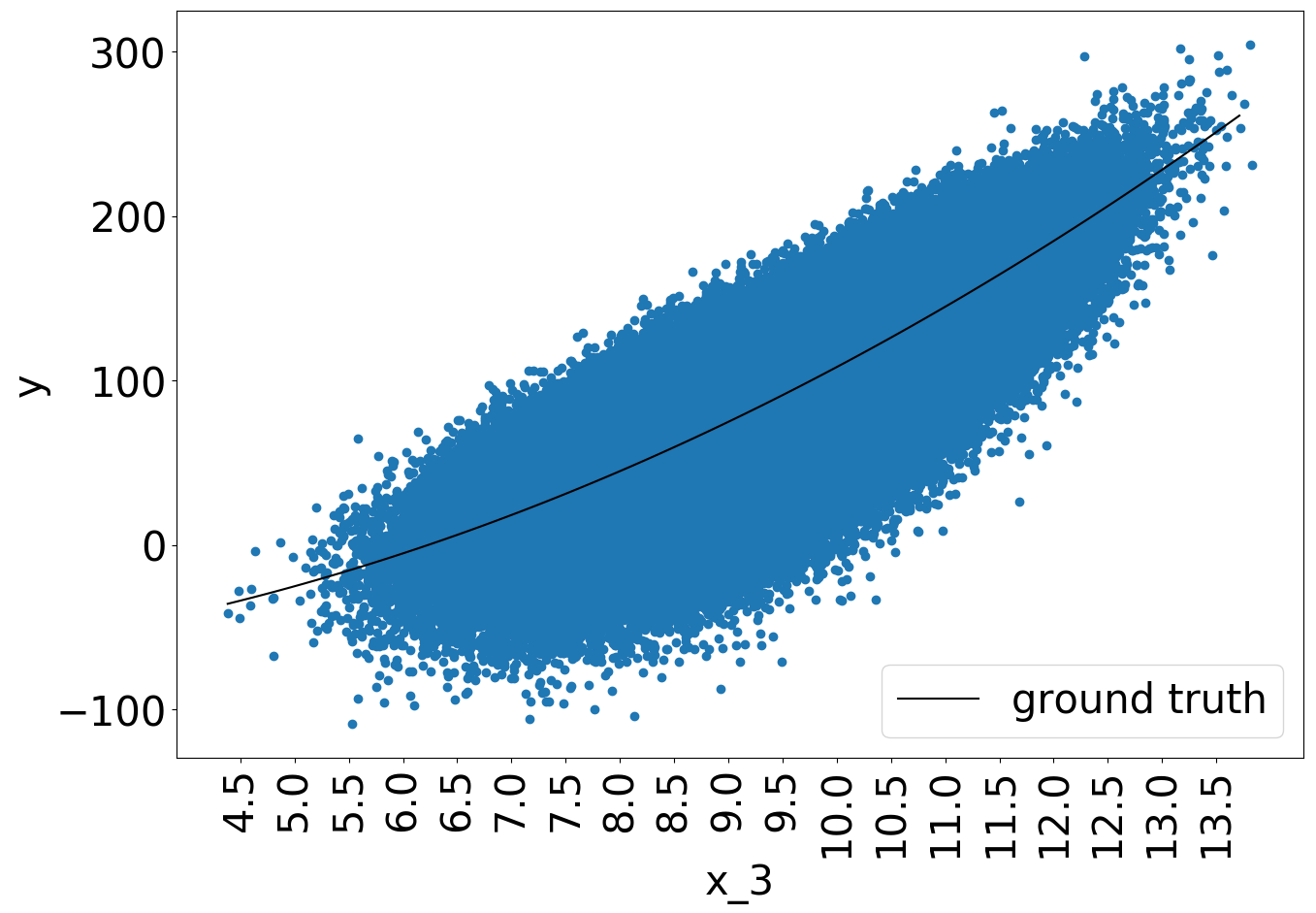}
\label{fig:scatter_2}}\\
\subfloat[]{\includegraphics[width=0.55\linewidth]{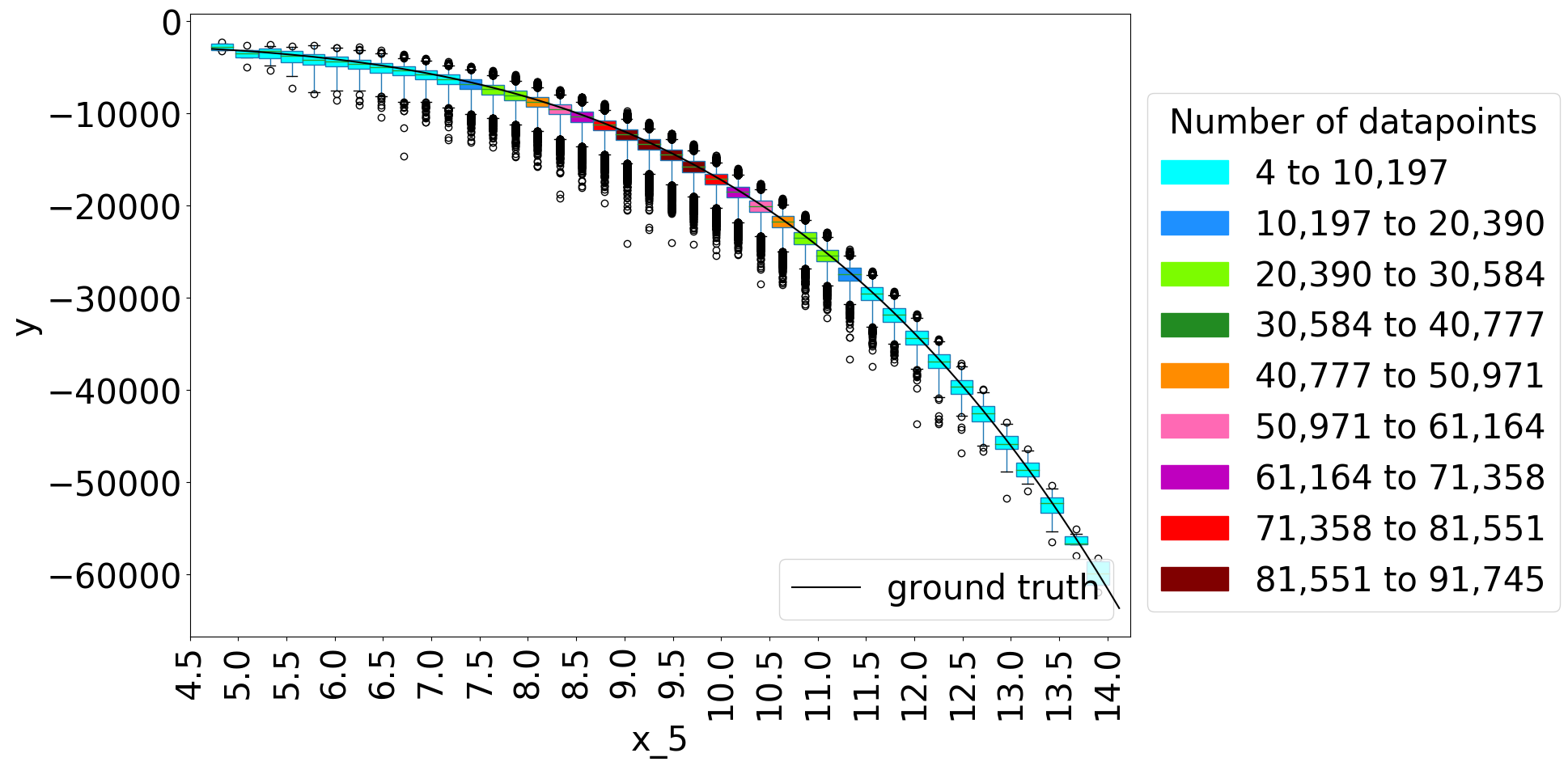}
\label{fig:box_4}}
\subfloat[]{\includegraphics[width=0.4\linewidth]{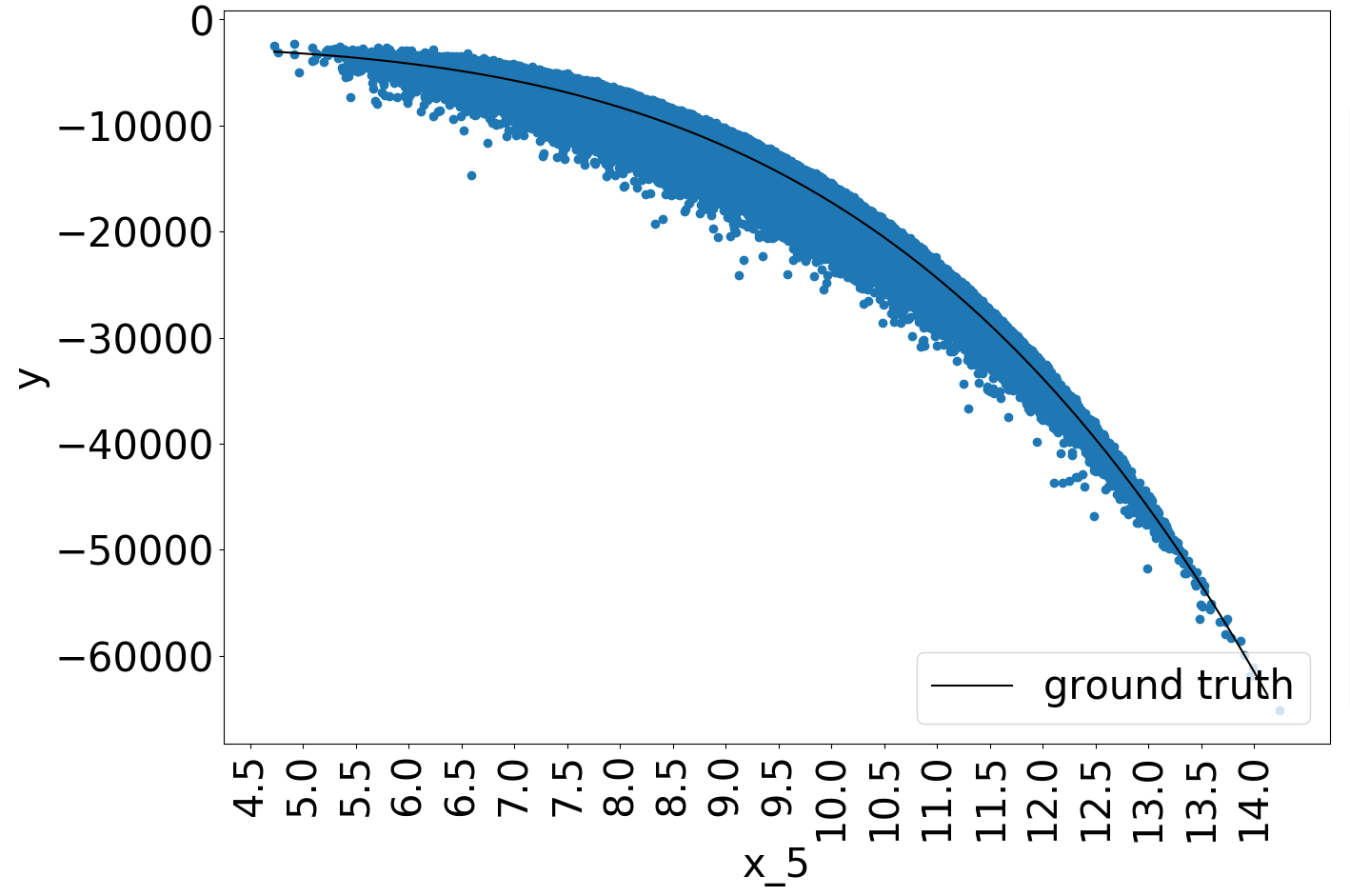}
\label{fig:scatter_4}}
\caption{Illustrations of different data sets generated for this study: 
A) boxplots showing the spread of y values for each interval of $x_{0}$ along with $f_{0}$ from a data set of \textbf{type a} from Table \ref{tab:data}, shading of each boxplot shows the quantity of datapoints in any given interval, illustrating the sparse areas of data created at the tails of the $x_{1}$ distribution; 
B) scatter plot of $x_{0}$ to $y$ for a data set of \textbf{type a} showing $f_{0}$;
C) boxplots showing the spread of y values for each interval of $x_{3}$ along with $f_{3}$ from a data set of \textbf{type h};
D) scatter plot of $x_{3}$ to $y$ for a data set of \textbf{type h} showing $f_{3}$;  
E) boxplots showing the spread of y values for each interval of $x_{5}$ along with $f_{5}$ from a data set of \textbf{type l};
F) scatter plot of $x_{5}$ to $y$ for a data set of \textbf{type l} showing $f_{5}$. 
 }\label{fig:data}
\end{figure}

This produces a data set for a regression where the known relationships between input and output variables are $\bm{W \cdot F}$ or $w_{i} \cdot f_{i}$ $\forall i$, which are all functions and will be denoted $\bm{f}_{i}$, and is the isolated relationship between the input $i$ and the output. A range of curvatures of $f$; noise in the $y$ direction; and covariance matrices for generating $X$ are tested, Table \ref{tab:data}, to identify the types of data where the error measure is effective.

\begin{table}[h]
\begin{center}
\begin{tabular}{ c|c|ccc| }
\multicolumn{2}{c}{} &\multicolumn{3}{c}{$f$} \\
\cline{3-5}
\multicolumn{1}{c}{}& &  $\mathds{P}_{0-1}$ & $\mathds{P}_{2-3}$ & $\mathds{P}_{4-5}$ \\ 
\cline{2-5}
\multirow{4}{*}{$\bm{\sigma}$}& 0.1 & \textbf{a}&\textbf{b}&\textbf{c} \\  
&1 &\textbf{d}&\textbf{e}&\textbf{f}\\    
&10&\textbf{g}&\textbf{h}&\textbf{i} \\
&100&\textbf{j}&\textbf{k}&\textbf{l}\\ 
\cline{2-5}
\end{tabular}
\end{center}\caption{Varieties of dataset generated for trialling the error measure, Illustrated in Figure \ref{fig:data}.}\label{tab:data}
\end{table}

The data generated for this study aims to replicate a scenario where a single target variable is dependent on 6 independent input variables, with their own specific relationship to the target. Each input variable has an arbitrarily generated polynomial relationship, $f_{i}$, with noise independent of the other input variables, the target, $y$, is the weighted average of each $f_{i}$.  

For data sets with non-linear polynomials this creates non-Gaussian noise in the $y$ distribution, violating assumption (ii). The reason for violating this assumption is that many applications require modelling of systems with non-Gaussian noise in the target variable. Although the noise introduced in these artificial data sets has the same Gaussian distribution for each input variable, their introduction at the input variable level is sufficient to provide non-Gaussian noise in the output variable. The interaction between input-level noise also produces heteroscedastic data sets, which violates assumption (iii).

The noise distributions are illustrated by the scatter plot in Figure \ref{fig:box_4} and \ref{fig:scatter_4}, the combination of noise from the other 5 inputs creates an increased spread below the concave $x_{5}-y$ relationship, explained by Jensen's inequality. This skewed conditional distribution of the output variable means that often the conditional median approximates the ground truth closer than the conditional mean.

Input variables are generated as normal distributions around an arbitrary mean, so there is not sufficient data across the entirety of each $x_{i}$ domain, this violates assumption (iv). The shading of boxplots in Figure  \ref{fig:box_0}, \ref{fig:box_2} and \ref{fig:box_4} illustrates the normal distribution of $x_{i}$ values in one of the generated data sets, where a mean of 9.25 and standard deviation of 1, has created ranges of sparse data for $x_{i} < 6$ and $x_{i} > 13$. 

This methodology produces a data set which is not equivalent to any specific data set, but allows validation of the different regression error measures on a data set that can be controlled whilst replicating the main features from real data. However, the artificial input values all have a normal distribution with equal mean and median values whereas many real-life situations include variables which show more complex distributions. Arbitrary regression data sets from the UCI Machine Learning Repository \cite{Dua2019}, as well as privately available data sets, were analysed to contextualise the artificial data sets used in this study. The input-output relationships in real data sets, such as wind turbine power, fish toxicity and housing price regression data, are seen to be similar to those used to generate the artificial data sets. 

\section{Assessment of the Mean Fit to Median Error Measure}\label{results}
Three separate data sets are tested for each type of data set in Table \ref{tab:data},  with different randomly generated parameters. For each of these data sets 1,500 neural networks are trained using normal backpropogation of random sizes ranging from (1,1) to (3,1000) and a loss function based on the Mean Absolute Error, full hyperparameters are provided in Table \ref{tab:newHyperParam}. After training, each network is tested with the newly derived Mean Fit to Median Error Measure, as well as Mean Absolute Error and Mean Squared Error, as these are the most popular error measures used for regression \cite{Fidles2007} \cite{Mccarthy2006}. The Mean Absolute  Error is used as the \mbox{Minkowski-r} metric to compare to because the curvature of the input-output relationships means the conditional median is closer to the ground truth than the conditional mean \cite{Bishop1995}, although the Mean Square Error was also trialled due to its prolific use in the field but showed similar behaviour to the Mean Absolute Error.

\begin{table}[h]
\begin{tabular}{c|c}
Hyperparameter & Value or set\\
\hline
\hline
Number of hidden layers & [1,3]\\
\hline
Number of neurons & \multirow{2}{*}{[1,1000]} \\ 
in each hidden layer & \\
\hline
Number of epochs & 20 \\
\hline
Batch Size & 50\\
\hline
Early Stopping Patience &  5 \\
\hline
Error function & Mean Absolute Error\\
\hline
Learning rule &  AdaMax \cite{Kingma2015} \\
\hline
Activation Function & ReLU\\
\hline
Regulariser & None\\
\hline
Initialiser & Random Normal ($\mu = 0, \sigma =0.1$)\\
\end{tabular}\centering{}\caption{Selected Hyperparameters}\label{tab:newHyperParam}
\centering
\end{table}

Since the data sets are all artificial, the underlying relationships within them are known explicitly, therefore it is possible to assess how well each network fits the ground truth of the data set. This measure is called `Mean Fit to the Ground Truth' which is derived similarly to the Mean Fit To Median with the difference that the input-output function $\phi$ is used to generate the data set,

\begin{equation}
        \text{Mean Fit to Ground Truth} = \frac{1}{m} \sum^{m}_{j=1} \frac{1}{n} \sum^{n}_{i=1} \left|\phi \left( \text{mid}(X^{j}_{i}), \text{med}(X^{k}) \text{ for } k\neq j \right) - p^{j}_{X_{i}}\right|.
        \label{eq:MAPE}
\end{equation}

\noindent This involves cycling from the minimum to maximum observed values, taking the true value of $\phi$ for the midpoint of each partition in the input domain, with all of the other inputs held at the median value. It is only possible to calculate this measure for artificial data sets.

The errors from 1,500 separate network runs are calculated and each error measure is normalised independently as the data sets are artificial, therefore the absolute magnitude of an error measure has limited meaning. The error measures are compared to assess if the Mean Fit to Median measure approximates the Mean Fit to the Ground Truth of the data set better than the Mean Absolute Error or Mean Squared Error. If there is a correlation between the Mean Fit to the Median and the Mean Fit to the Ground Truth then the Mean Fit to the Median can be used as a measure that acts as a proxy for how well a network approximates the ground truth of these data sets.

\begin{figure}[h!]
\centering
\subfloat[]{\includegraphics[width=0.46\linewidth]{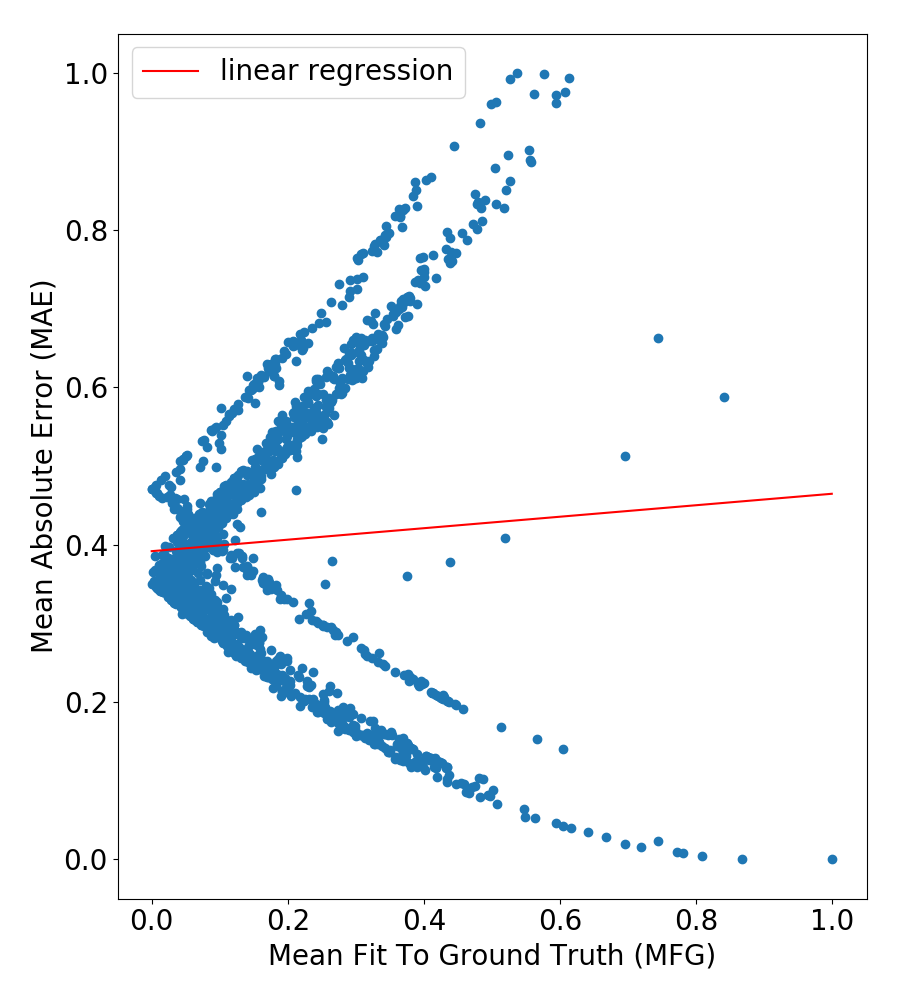}
\label{fig:0-2_accuracy}}
\subfloat[]{\includegraphics[width=0.46\linewidth]{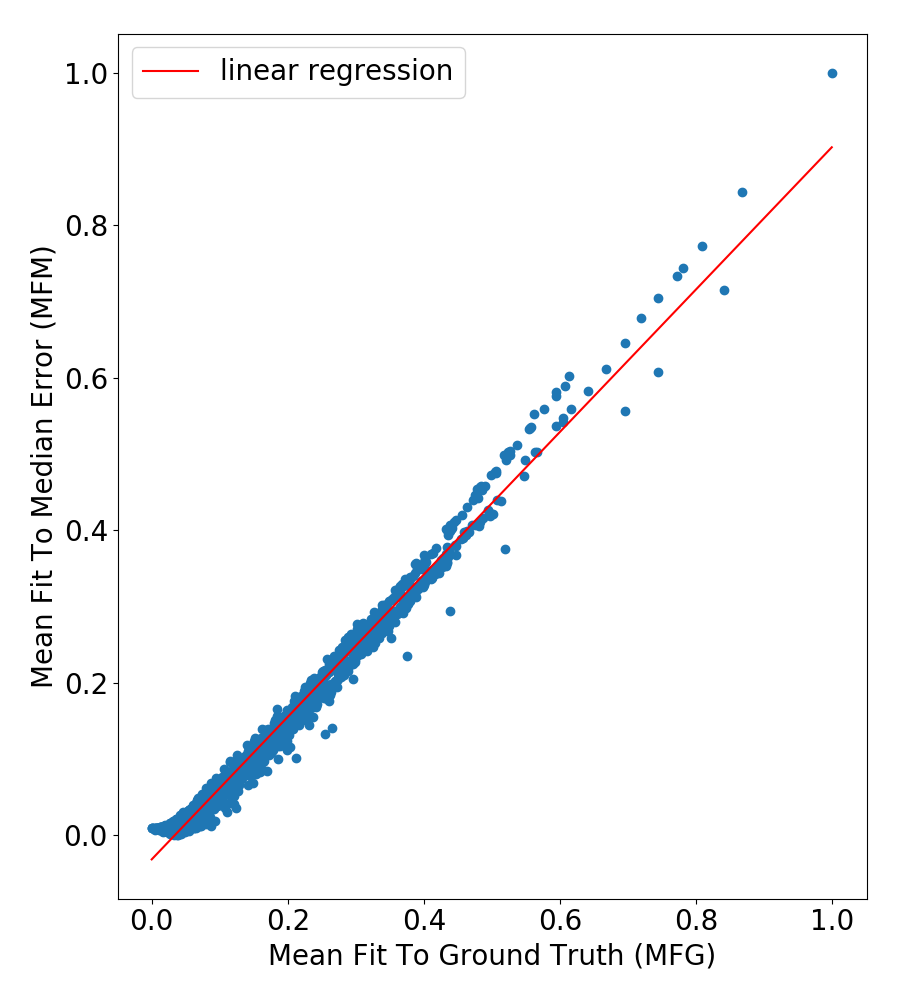}
\label{fig:0-2_median}}
\caption{1,500 networks assessed on the data set of \textbf{type d}, where the data is fit with a linear regression a) the Mean Absolute Error against Mean Fit To Ground Truth with $R^{2}$ of 0.0036 and b) the Mean Fit To Median against Mean Fit To Ground Truth with $R^{2}$ of 0.9801.}\label{fig:0-2}
\end{figure}

\begin{figure}[h!]
\centering
\subfloat[]{\includegraphics[width=0.46\linewidth]{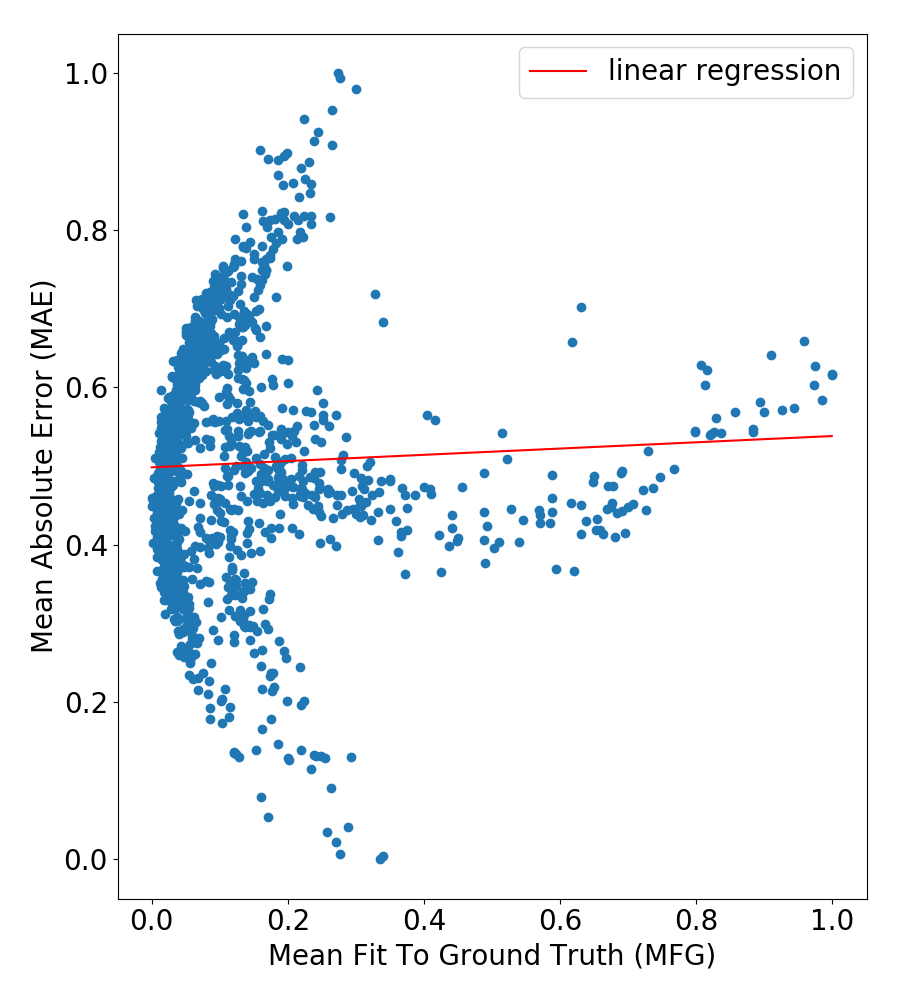}
\label{fig:2-4_accuracy}}
\hfil
\subfloat[]{\includegraphics[width=0.46\linewidth]{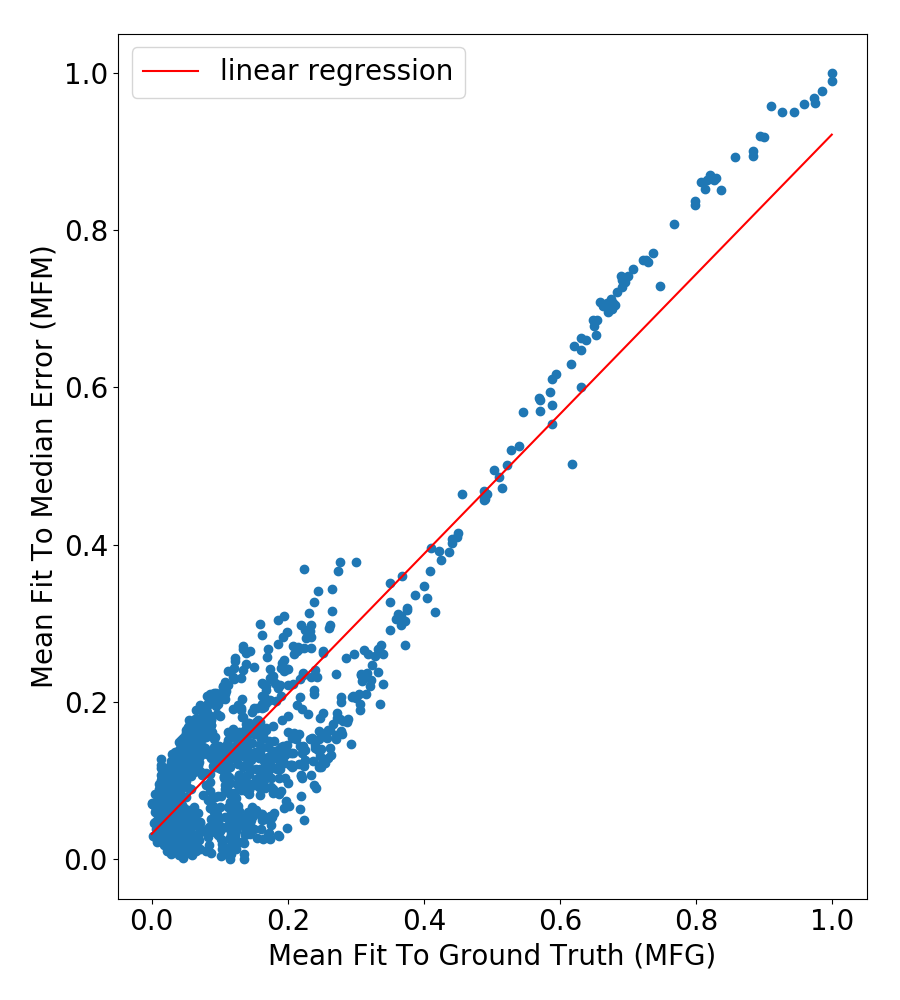}
\label{fig:2-4_median}}
\caption{1,500 networks assessed on the data set of \textbf{type b}, where the data is fit with a linear regression a) the Mean Absolute Error against Mean Fit To Ground Truth with $R^{2}$ of 0.0016 and b) the Mean Fit To Median against Mean Fit To Ground Truth with $R^{2}$ of 0.8464.}\label{fig:2-4}
\end{figure}

\begin{figure}[h!]
\centering
\subfloat[]{\includegraphics[width=0.46\linewidth]{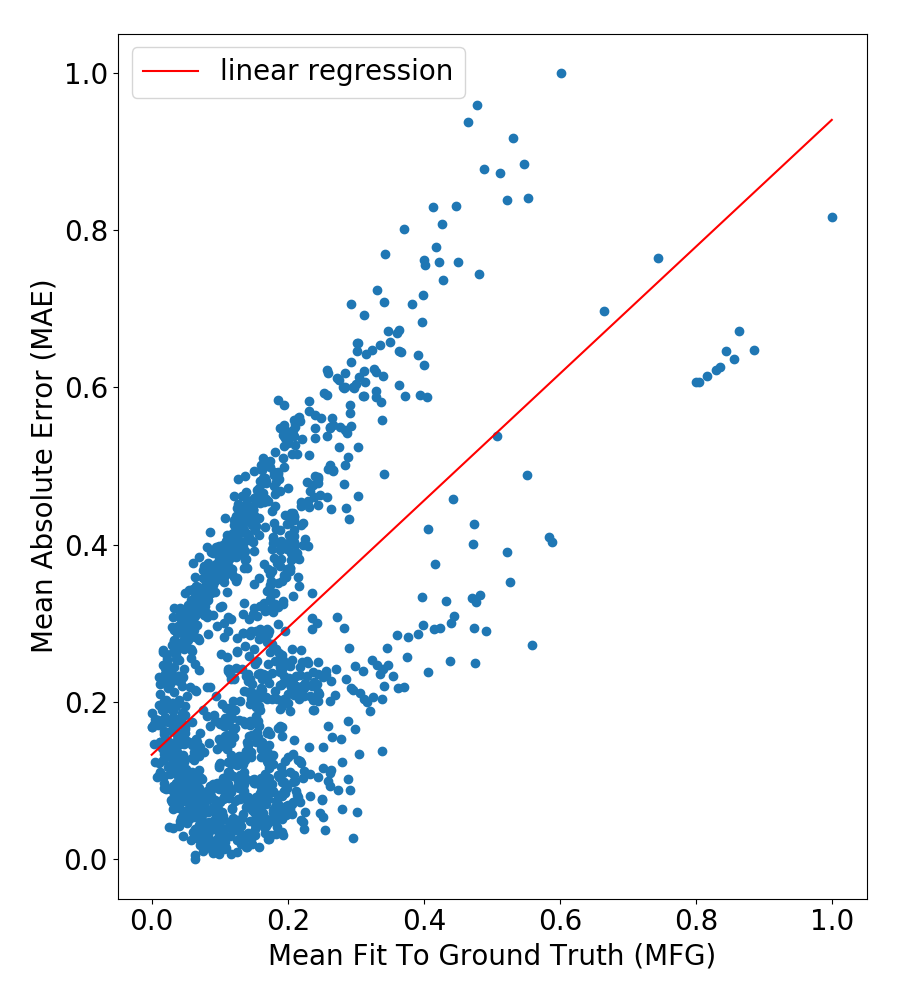}
\label{fig:4-6_accuracy}}
\hfil
\subfloat[]{\includegraphics[width=0.46\linewidth]{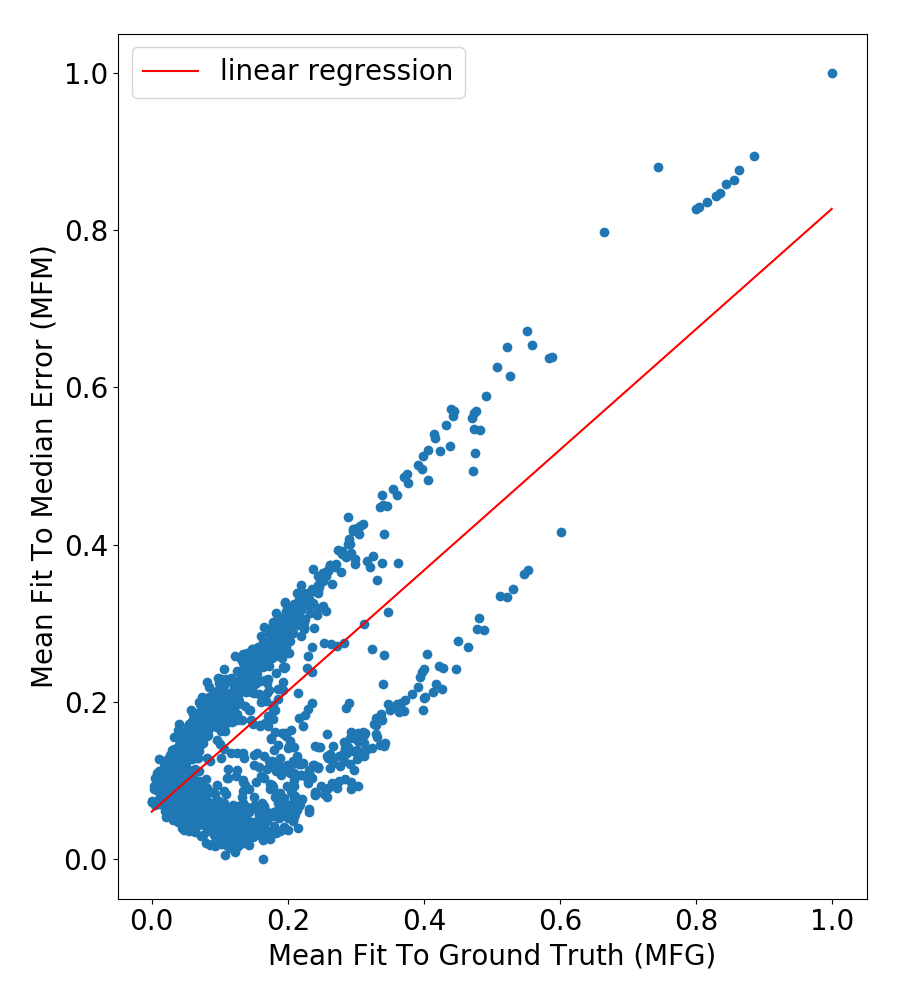}
\label{fig:4-6_median}}
\caption{1,500 networks assessed on the data set of \textbf{type l}, where the data is fit with a linear regression a) the Mean Absolute Error against Mean Fit To Ground Truth with $R^{2}$ of 0.2704 and b) the Mean Fit To Median against Mean Fit To Ground Truth with $R^{2}$ of 0.5184.}\label{fig:4-6}
\end{figure}

The relationships between the Mean Absolute Error and the Mean Fit to the Ground Truth show a high variation across the different data sets, nearly all relationships involve bifurcations, although this becomes less prominent as the degree of the polynomials in the data set decreases. Figure \ref{fig:0-2_accuracy} shows an almost perpendicular bifurcation at 0.4 normalised Mean Absolute Error whereas Figures \ref{fig:2-4_accuracy} and \ref{fig:4-6_accuracy} have a more irregular pattern but still show prominent bifurcations. The cause of these patterns is discussed in Section \ref{discussion}. Contrastingly, the Mean Fit to the Median and the Mean Fit to the Ground Truth clearly approach more of a one-to-one relationship. This is quantified using $R^{2}$ from a linear regression performed on the normalised data, the lines are shown on Figures \ref{fig:0-2}, \ref{fig:2-4} and \ref{fig:4-6}.

\begin{figure}[h!]
\centering
\subfloat[]{\includegraphics[width=0.46\linewidth]{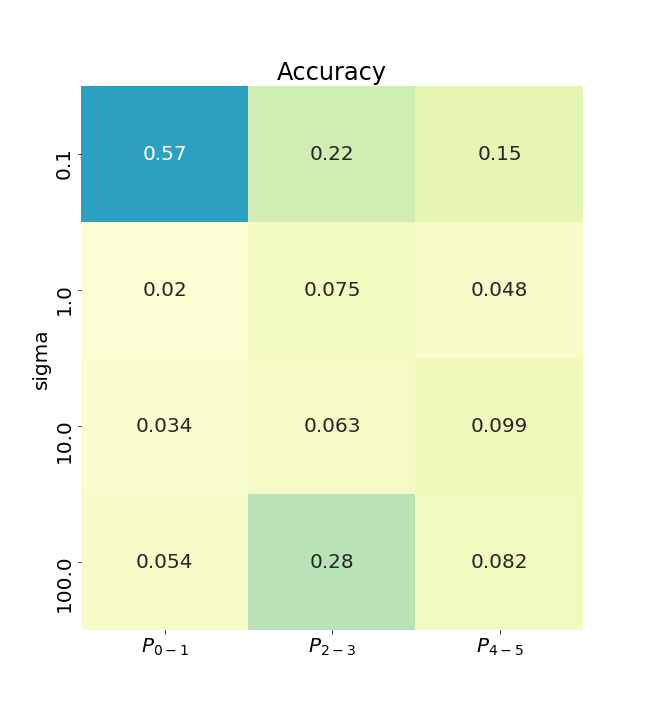}
\label{fig:heatmap_accuracy}}
\hfil
\subfloat[]{\includegraphics[width=0.46\linewidth]{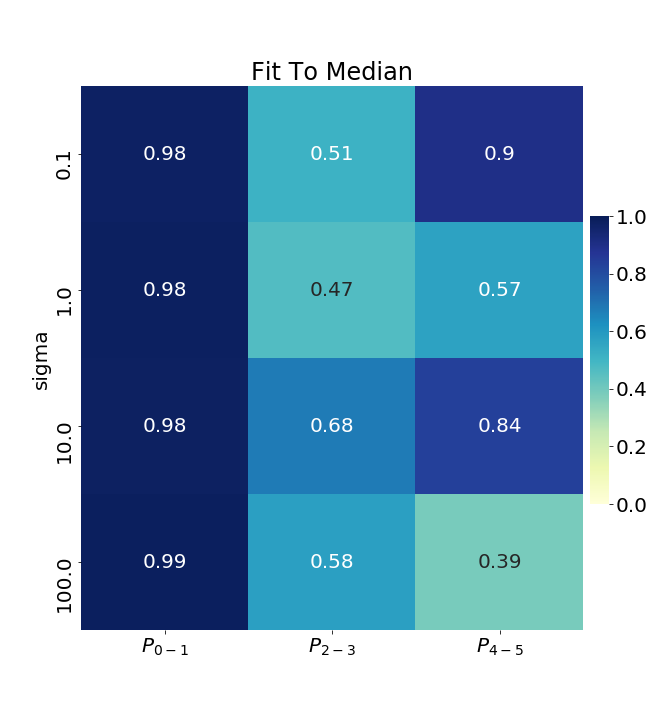}
\label{fig:heatmap_median}}
\caption{Heatmaps illustrating the average $R^{2}$ between error measures and the Mean Fit To Ground Truth from the different types of data set, stipulated in Table \ref{tab:data} a) the average $R^{2}$ of the traditional, Mean Absolute Error, error measure against the Mean Fit To Ground Truth and b) the average $R^{2}$ of the proposed, Mean Fit To Median, error measure against the Mean Fit To Ground Truth.}\label{fig:heatmaps}
\end{figure}

For a data set with entirely linear $f_{i}$ and a low standard deviation of noise, there is almost total agreement between the Mean Fit to the Median and Mean Fit to the Ground Truth, Figure \ref{fig:0-2_median}. This correlation lessens as the standard deviation of noise increases and as the $f_{i}$ increases in degree, or curvature.  Figure \ref{fig:2-4_median} shows results from a data set with increased curvature but low noise, compared to Figure  \ref{fig:0-2_median}, and although the correlation is still prominent, the $R^{2}$ has decreased from 0.98 to 0.85. The data sets with polynomial $f_{i}$ of degrees 4-5 and high noise levels are hard to model, requiring at least three hidden layers in a neural network to approximate accurately. The results from this data set show the lowest correlation between Mean Fit to the Median and Mean Fit to the Ground Truth with an $R^{2}$ of 0.52, but this is still significantly higher than the correlation between Mean Absolute Error and Mean Fit to the Ground Truth.

This decrease in correlation between Mean Fit to the Median and Mean Fit to the Ground Truth as more noise and curvature is introduced to the data sets can be explained by Jensen's inequality. The increase in noise and curvature of $f_{i}$ create a larger skew in the conditional distributions of the output variable, this skew causes the mean and median of the conditional distribution to move away from the ground truth or $\phi(X_{i})$. If the median is not an accurate representation of the ground truth then the Mean Fit to the Median will not be as effective as it directly measures the distance of a prediction from the conditional median. However, the Mean Fit to the Median approximates the Mean Fit to the Ground Truth better than the Mean Absolute Error for all analysed data sets. Networks with low Mean Fit to the Median therefore replicate the ground truth better than networks with low Mean Absolute Error.

The $R^{2}$ values from all of the regressions on the 36 data sets are collated and this is illustrated in Figure \ref{fig:heatmap_accuracy} and \ref{fig:heatmap_median}. Comparing the average $R^{2}$ value of the linear regressions performed for each different type of data set, shows where the Mean Fit to the Median is most effective. All of the $R^{2}$ values for the Mean Absolute Error are lower, 0.02-0.57, than their Mean Fit to the Median counterparts, 0.39-0.99, from Figure \ref{fig:heatmap_accuracy}, where a higher $R^{2}$ indicates a tighter positive correlation between the error measure and the Mean Fit to the Ground Truth of the data set. For data sets with lower degree polynomial relationships the $R^{2}$ values for the Mean Fit to the Median are particularly high, approaching 1, Figure \ref{fig:heatmap_median}. However, data sets with these characteristics should be simple enough to analyse that advanced machine learning techniques, such as neural networks, do not produce any further insight than simpler analysis techniques.

All of the $R^{2}$ values for the Mean Fit to the Median are higher than the Mean Absolute Error values, meaning the Fit to Median is better at assessing how well a network models then ground truth than the Mean Absolute Error. This increase in $R^{2}$ ranges from 0.29-0.96, with an average of 0.6, across all 36 data sets. Choosing networks with a low Mean Fit to the Median increases the likelihood of a better model of the underlying relationships within a data set, when compared to relying on the most commonly used error measures.

\section{Investigation into the Poor Bias from Minkowski-r Error Measures}\label{discussion}

\begin{figure}[h!]
\centering
\subfloat[]{\includegraphics[width=0.41\linewidth]{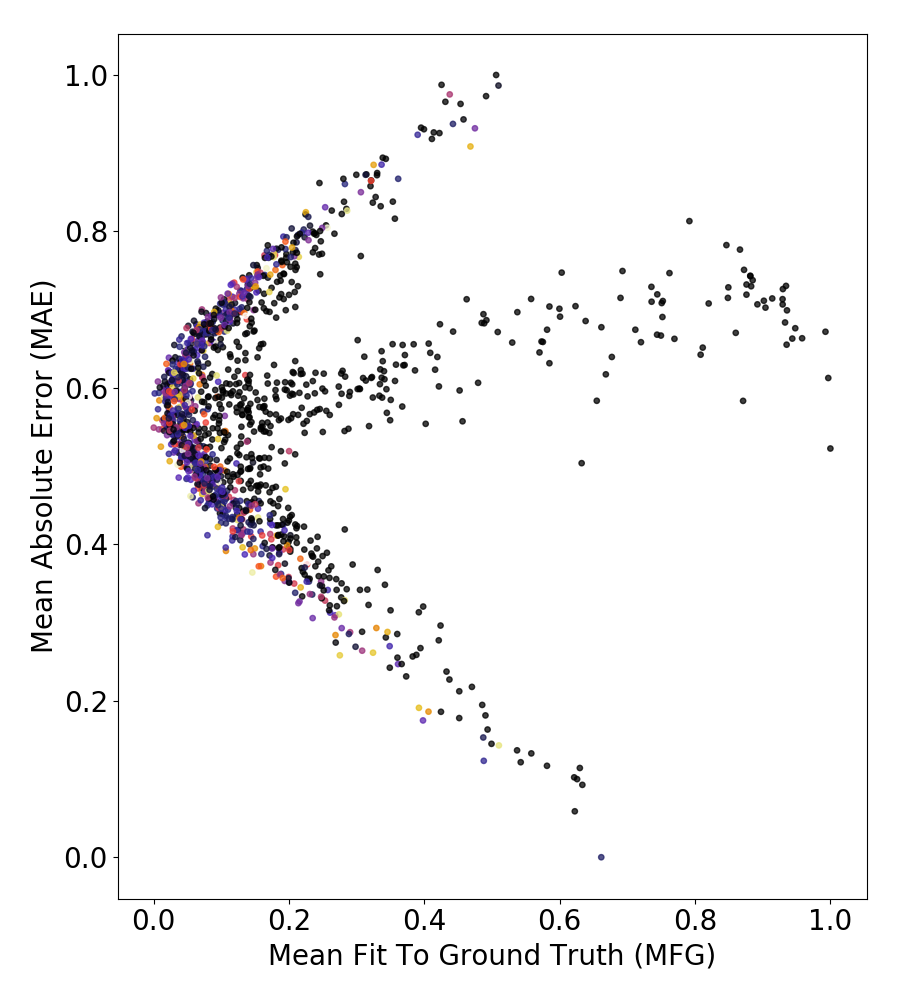}
\label{fig:complexity_accuracy}}
\hfil
\subfloat[]{\includegraphics[width=0.5\linewidth]{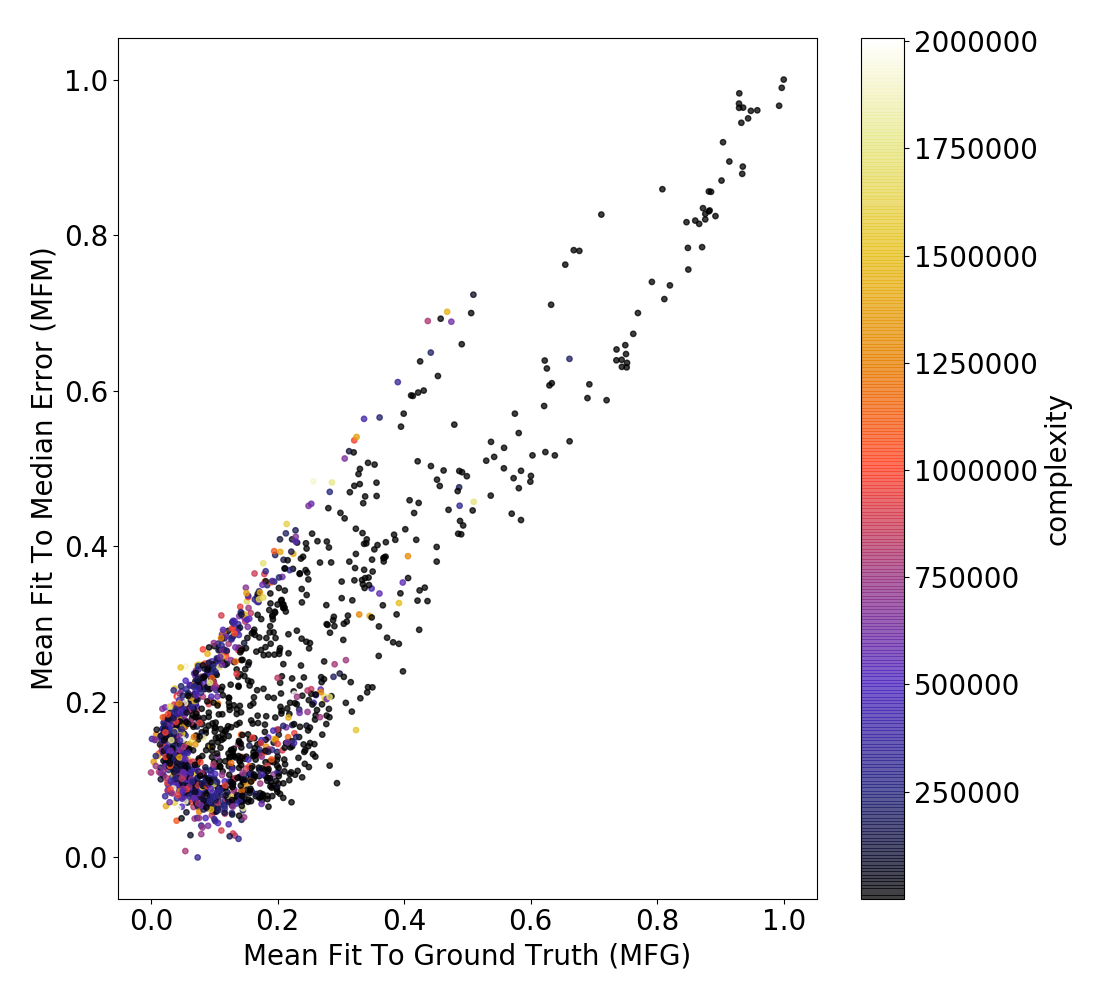}
\label{fig:complexity_median}}
\caption{1,500 networks assessed on the data set of \textbf{type e} with a) Mean Absolute Error against Mean Fit To Ground Truth and b) Mean Fit To Median against Mean Fit To Ground Truth. Where complexity, or number of connections in each network, is indicated by the colour of each point and shows no pattern for either plot.}\label{fig:complexity}
\end{figure}

The causes of the bifurcations in the relationship between Mean Absolute Error and the Mean Fit to the Ground Truth are investigated. A major factor affecting the error of a trained network is its size and shape. The networks trained had a random number of layers and neurons, with a maximum size of (3,1000) and therefore the bifurcations imply that some networks trained with Mean Absolute Error have a bias to the ground truth and that some have a bias to a pattern that is not the ground truth, negatively affecting their ability to replicate this relationship. One of the assumptions required for minimum Mean Absolute Error to model the conditional median of a data set is the use of a sufficiently large neural network. As randomly sized networks are used in this study this assumption is not necessarily satisfied. Therefore, to evaluate the effect of the size of a network on the error, the complexity, or number of connections, are indicated by colour in Figure \ref{fig:complexity}.

There is no clear relationship between the size of the network and error for either the Mean Absolute Error, Figure \ref{fig:complexity_accuracy}, the Mean Fit to the Median or the Mean Fit to the Ground Truth, Figure \ref{fig:complexity_median}. This suggests that the bifurcations are not caused by over or under-fitting of the networks. Although some relationship between the network size and error would be expected when using a large range of network sizes, the maximum number of layers trialled is only 3 and as all of the networks employ early stopping it is suggested this procedure decreases the chances of networks overfitting.

It is confirmed that, on the whole, networks with low Mean Fit To Median also have a low Mean Absolute Error but not the lowest observed Mean Absolute Error, so fitting to the ground truth of a data set does not produce the lowest point-to-point accuracy but does provide an acceptable level. For all 36 trialled data sets, 70\% show lower Mean Absolute Error as Mean Fit To Median lowers, this means the Mean Fit To Median error measure can be used alongside Mean Absolute Error to produce networks which do not compromise the point-to-point accuracy of prediction significantly, but do increase the fit to the ground truth of the data set.

\begin{figure}[h!]
\centering
\subfloat[]{\includegraphics[width=0.41\linewidth]{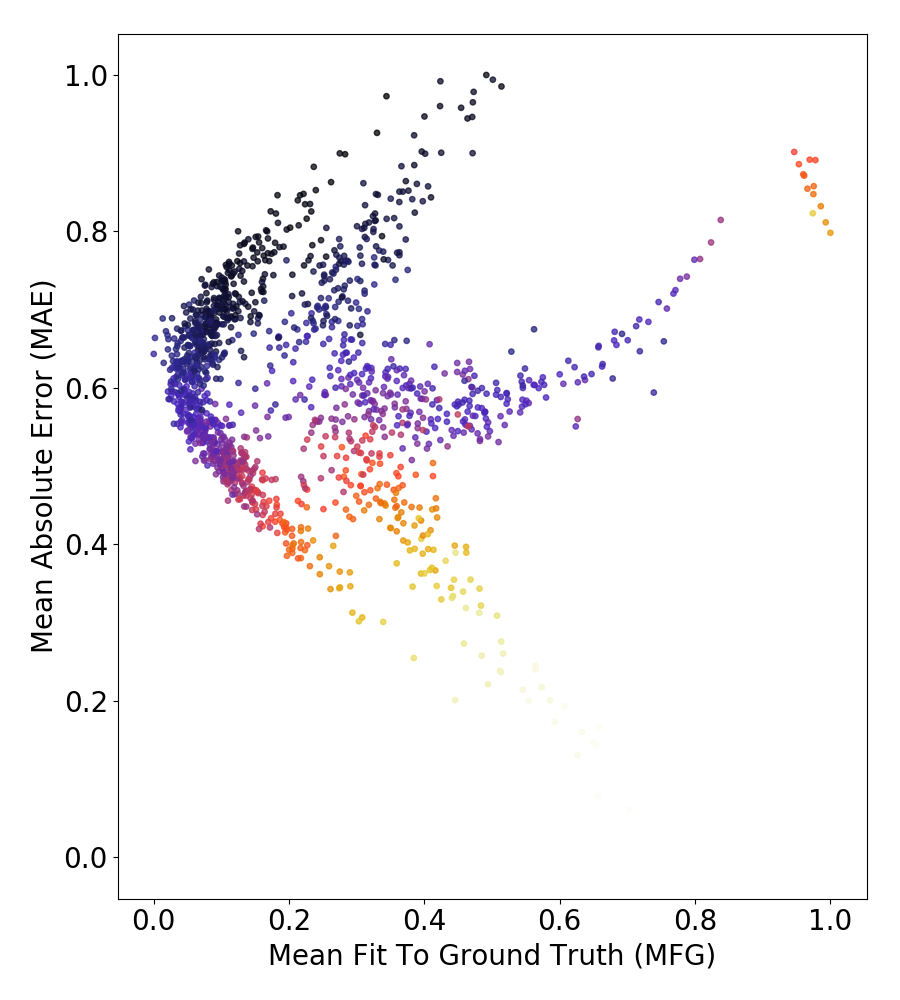}\label{fig:mean-median_accuracy}}
\hfil
\subfloat[]{\includegraphics[width=0.5\linewidth]{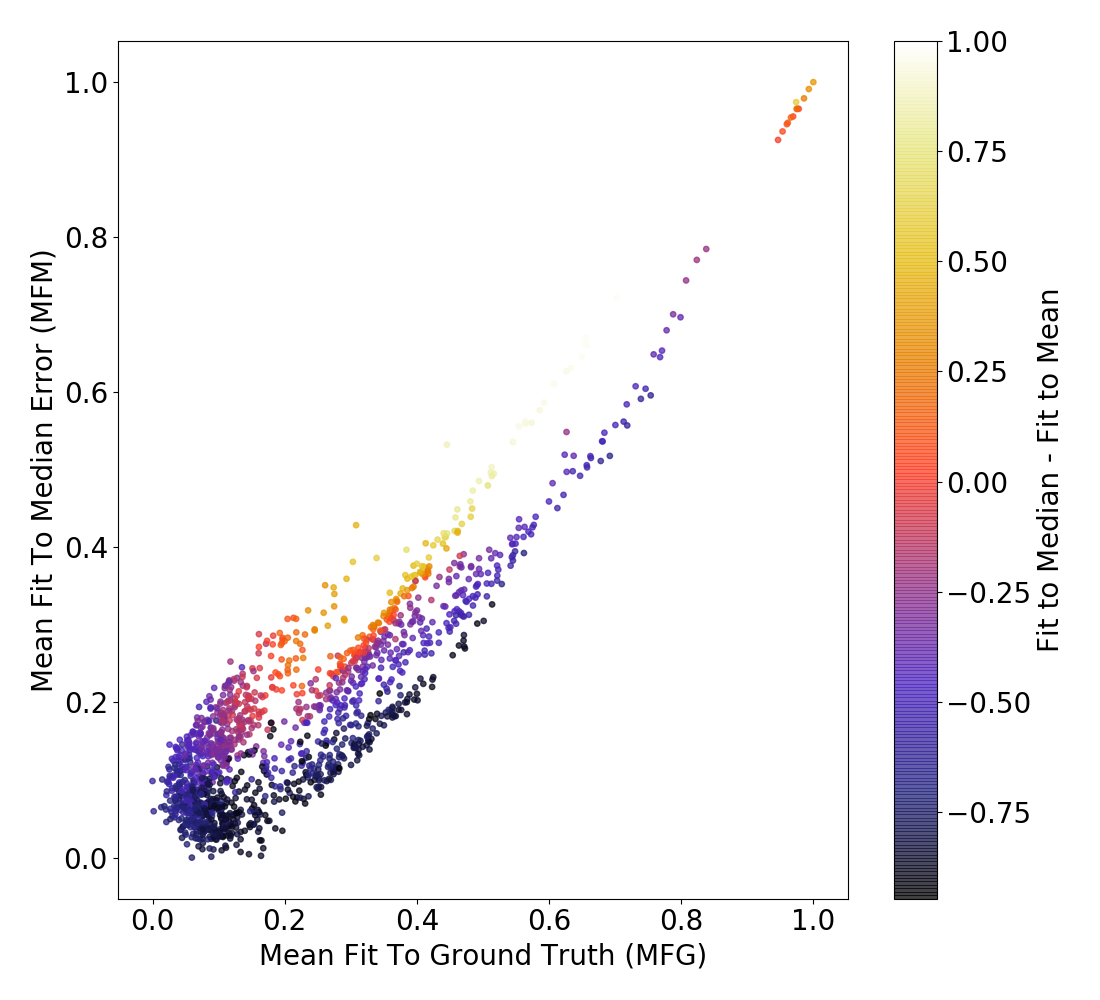}\label{fig:mean-median_median}}
\caption{1,500 networks assessed on the data set of \textbf{type l} with a) Mean Absolute Error against Mean Fit To Ground Truth and b) Mean Fit To Median against Mean Fit To Ground Truth. The difference between the Mean Fit To Median and `Fit to Mean' of each network is indicated by their colour, with networks where the `Fit to Mean' is lower coloured lighter and networks where the Mean Fit To Median is lower coloured darker.}
\label{fig:mean-median}
\end{figure}

The characteristics of the data sets, their distributions and underlying relationships, are investigated as possible causes of the bifurcations. The normalised difference between the Mean Fit to the Median and `Fit to Mean' error measures are indicated by data point colour in Figure \ref{fig:mean-median}. The `Fit to Mean' error measure is derived similarly to the Mean Fit to the Median except the proxy curve is $E[\phi(X)]$ as opposed to $med(\phi(X))$ in Equation \ref{eq:MFTM}.

The networks with the lowest Mean Absolute Error approximate relationships which are closer to the mean of the data than the median of the data and hence have a positive difference between Mean Fit to the Median and `Fit to Mean'. These lighter points in Figure \ref{fig:mean-median_accuracy} form the bottom tail of the bifurcation and have a higher Mean Fit to the Ground Truth error than the main body. This suggests that as networks more closely fit to the conditional mean of a data set, this approximation stops fitting to the ground truth. As the difference between Mean Fit to the Median and `Fit to Mean’ becomes negative the networks fits closer to the conditional median of the data set than the mean, indicated by the darker data points. The networks which have the lowest Mean Fit to the Median, comparative to the `Fit to Mean’, diverge away from the main body as the upper part of the bifurcation as the Mean Absolute Error from these networks is high.

Therefore, the bifurcations noted for these data sets may be caused by the bias created by training with the Mean Absolute Error. The minimum Mean Absolute Error does not guarantee a fit of the conditional median as shown in previous sections, in this particular data set the networks with lowest Mean Absolute Error are closest to the conditional mean of the data set. For the data sets used in this study the conditional median is a better approximation of the ground truth than the conditional mean, hence the lower tail in Figure \ref{fig:mean-median_accuracy}. which contains networks which fit closely to the conditional mean but do not closely fit to the ground truth. This shows an unhelpful inductive bias is created by the use of the Mean Absolute Error. 

New error measures are required to improve regression  approximations of the ground truth of a data set, these will need to take into account the origins of the unhelpful bias created by Minkowski-r error measures. As neural networks display low neuroplasticity, the bias will be decided early on in the training process which makes it difficult to overcome using conventional error measures. This suggests that the bifurcation of networks into helpful or unhelpful bias originates from inherent stochasticity in the training of a neural network. Explicitly this is a combination of the initialisation and the order the data set is shown to the network in the first epoch, as well as any stochastic elements introduced by learning rules. 

\section{Conclusion}

Regression methods can report low conventional error measures by mapping arbitrary patterns between the inputs and outputs. In many applications, such as where the data is non-linear or sparse, methods trained on these error measures demonstrate a poor approximation of the ground truth, which reduces trust in the method. This paper develops a error measure, the Mean Fit To Median error, which improves the approximation of the conditional average for applications where no understanding of the input-output relationships exist but where a better modelling of the ground truth would provide a greater understanding of the problem and allow for extrapolation. The error measure is robust across a larger range of data sets than the traditional error measures and can be used in a regression method to better approximate the ground truth, in order to provide more explainable machine learning. The Mean Fit To Median error, which measures how far from a proxy of the conditional median predictions are, is compared to the traditional Minkowski-r error measures on 36 different data sets and is shown to have a stronger correlation to the Fit to Ground Truth, with an average increase in $R^{2}$ of 0.6 compared to the Mean Absolute Error. Methods with low Mean Fit To Median error model the ground truth of a data set more reliably than those with a low Mean Absolute Error, by creating a bias to the ground truth.  Selecting trained methods with a low Fit To Median Error, in combination with the use of traditional loss functions, increases the likelihood that the relationships between the inputs and the outputs of a regression problem are accurately mapped, even when there is no understanding of what these relationships are.  

\section*{Acknowledgments}
The authors thank Shell Shipping and Maritime for funding this work. The work was also kindly supported by the Lloyds Register Foundation. The authors acknowledge the use of the IRIDIS High Performance Computing Facility, and associated support services at the University of Southampton, in the completion of this work. 

\printbibliography

\end{document}